\documentclass[compsoc,journal]{IEEEtran}

\usepackage{amsmath,amsfonts}
\usepackage{algorithmic}
\usepackage{algorithm}
\usepackage{array}
\usepackage{textcomp}
\usepackage{stfloats}
\usepackage{url}
\usepackage{verbatim}

\usepackage{graphicx}
\usepackage{caption}
\usepackage{subcaption}
\usepackage{colortbl}
\usepackage{transparent}

\usepackage{cite}

\usepackage{hyperref}

\usepackage{listings}
\usepackage{xcolor}

\usepackage{adjustbox}

\usepackage[acronym]{glossaries}

\definecolor{codegreen}{rgb}{0,0.6,0}
\definecolor{codegray}{rgb}{0.5,0.5,0.5}
\definecolor{codepurple}{rgb}{0.58,0,0.82}
\definecolor{backcolour}{rgb}{0.95,0.95,0.92}
\definecolor{spiker_green}{rgb}{0,0.4,0.4}
\definecolor{spiker_green}{rgb}{0,0.4,0.4}
\definecolor{mnist_blue}{rgb}{0.153, 0.251, 0.545}
\definecolor{shd_red}{rgb}{0.803, 0.149, 0.149}

\lstdefinestyle{mystyle}{
  backgroundcolor=\color{backcolour}, commentstyle=\color{codegreen},
  keywordstyle=\color{magenta},
  numberstyle=\tiny\color{codegray},
  stringstyle=\color{codepurple},
  basicstyle=\ttfamily\footnotesize,
  breakatwhitespace=false,         
  breaklines=true,                 
  captionpos=b,                    
  keepspaces=true,                 
  numbers=left,                    
  numbersep=5pt,                  
  showspaces=false,                
  showstringspaces=false,
  showtabs=false,                  
  tabsize=2
}

\newcommand{\ignore}[1]{}

\glsdisablehyper
\newacronym{ai}{AI}{Artificial Intelligence}
\newacronym{if}{IF}{Integrate and Fire}
\newacronym{iot}{IoT}{Internet of Things}
\newacronym{ann}{ANN}{Artificial Neural Network}
\newacronym{snn}{SNN}{Spiking Neural Network}
\newacronym{rsnn}{RSNN}{Recurrent Spiking Neural Network}
\newacronym{dnn}{DNN}{Deep Neural Network}
\newacronym{cnn}{CNN}{Convolutional Neural Network}
\newacronym{lif}{LIF}{Leaky Integrate and Fire}
\newacronym{asic}{ASIC}{Application-Specific Integrated Circuit}
\newacronym{fpga}{FPGA}{Field Programmable Gate Array}
\newacronym{stdp}{STDP}{Spike-Timing-Dependent Plasticity}
\newacronym{wta}{WTA}{Winner Takes All}
\newacronym{rtl}{RTL}{Register Transfer Level}
\newacronym{hdl}{HDL}{Hardware Description Language}
\newacronym{vhdl}{VHDL}{VHSIC Hardware Description Language}
\newacronym{sota}{SOTA}{State Of The Art}
\newacronym{sbs}{SbS}{Spike-by-Spike}
\newacronym{nn}{NN}{Neural Network}
\newacronym{ml}{ML}{Machine Learning}
\newacronym{scnn}{SCNN}{Spiking Convolutional Neural Networks}
\newacronym{fc}{FC}{Fully-Connected}
\newacronym{fffc}{FF-FC}{Feed-Forward Fully-Connected}
\newacronym{fcr}{FC-R}{Fully-Connected Recurrent}
\newacronym{rcr}{RC-R}{Randomly-Connected Recurrent}
\newacronym{fsm}{FSM}{Finite State Machine}
\newacronym{pu}{PU}{Processing Unit}
\newacronym{cpu}{CPU}{Central Processing Unit}
\newacronym{gpu}{GPU}{Graphic Processing Unit}
\newacronym{aer}{AER}{Address Event Representation}
\newacronym{ini}{INI}{Institute of Neuro-Informatics}
\newacronym{eth}{ETH}{Eidgenössische Technische Hochschule}
\newacronym{dse}{DSE}{Design Space Exploration}
\newacronym{soc}{SoC}{System on Chip}
\newacronym{MAC}{MAC}{Multiply and Accumulate}
\newacronym{BRAM}{BRAM}{Block RAM}
\newacronym{SRAM}{SRAM}{Static Random Access Memory}
\newacronym{DRAM}{DRAM}{Dynamic Random Access Memory}
\newacronym{LUT}{LUT}{Look Up Table}
\newacronym{FF}{FF}{Flip Flop}
\newacronym{DVS}{DVS}{Dynamic Vision Sensor}
\newacronym{SHD}{SHD}{Spiking Heidelberg Dataset}
\newacronym{BPTT}{BPTT}{Back-Propagation Through Time}
\newacronym{CU}{CU}{Control Unit}
\newacronym{ROM}{ROM}{Read Only Memory}
\newacronym{RAM}{RAM}{Random Access Memory}
\newacronym{DP}{DP}{Data Path}
\newacronym{GPU}{GPGPU}{General Purpose Graphic Processing Unit}
\newacronym{TPU}{TPU}{Tensor Processing Unit}
\newacronym{SIMD}{SIMD}{Single Instruction Multiple Data}
\newacronym{NLP}{NLP}{Natural Language Processing}

\newcolumntype{?}{!{\vrule width 1pt}}
\newcolumntype{m}{>{\columncolor{mnist_blue!30}}c}
\newcolumntype{s}{>{\columncolor{shd_red!30}}c}

\begin{document}

	\newcommand{\spikerframework}{Spiker+}

    \IEEEtitleabstractindextext{
	\begin{abstract}
	\glsresetall
	Including \glspl{ann} in embedded systems at the edge allows applications to exploit \gls{ai} capabilities directly within devices operating at the network periphery, facilitating real-time decision-making. Especially critical in domains such as autonomous vehicles, industrial automation, and healthcare, the use of \glspl{ann} can enable these systems to process substantial data volumes locally, thereby reducing latency and power consumption. Moreover, it enhances privacy and security by containing sensitive data within the confines of the edge device.
	The adoption of \glspl{snn} in these environments offers a promising computing paradigm, mimicking the behavior of biological neurons and efficiently handling dynamic, time-sensitive data. However, deploying efficient \glspl{snn} in resource-constrained edge environments requires hardware accelerators, such as solutions based on \glspl{fpga}, that provide high parallelism and reconfigurability.
	This paper introduces \spikerframework{}, a comprehensive framework for generating efficient, low-power, and low-area customized \glspl{snn} accelerators on \glspl{fpga} for inference at the edge. \spikerframework{} presents a configurable multi-layer hardware \glspl{snn}, a library of highly efficient neuron architectures, and a design framework, enabling the development of complex neural network accelerators with few lines of Python code. \spikerframework{} is tested on two benchmark datasets, the MNIST and the \gls{SHD}. On the MNIST, it demonstrates competitive performance compared to state-of-the-art \gls{snn} accelerators. It outperforms them in terms of resource allocation, with a requirement of 7,612 logic cells and 18 \glspl{BRAM}, which makes it fit in very small \glspl{fpga}, and power consumption, draining only 180mW for a complete inference on an input image. The latency is comparable to the ones observed in the state-of-the-art, with 780$\mu$s/img. To the authors' knowledge, \spikerframework{} is the first \glspl{snn} accelerator tested on the \gls{SHD}. In this case, the accelerator requires 18,268 logic cells and 51 \glspl{BRAM}, with an overall power consumption of 430mW and a latency of 54 $\mu$s for a complete inference on input data. This underscores the significance of \spikerframework{} in the hardware-accelerated \gls{snn} landscape, making it an excellent solution to deploy configurable and tunable \gls{snn} architectures in resource and power-constrained edge applications.
	\glsresetall
	\end{abstract}

	\begin{IEEEkeywords}
	    Spiking Neural Networks, LIF, FPGA, Neuromorphic accelerator, Edge computing, Artificial Intelligence, Frugal AI.
	\end{IEEEkeywords}

	}

	\title{\spikerframework: a framework for the generation of efficient Spiking Neural Networks FPGA accelerators for inference at the edge}

	\author{
		Alessio Carpegna,~\IEEEmembership{Student memeber,~IEEE}, 
		Alessandro Savino, ~\IEEEmembership{Senior member,~IEEE},
		Stefano Di Carlo, ~\IEEEmembership{Senior member,~IEEE}
		\thanks{
			Alessio Carpegna, Alessandro Savino, and Stefano Di Carlo are with the Department of Control and Computer Engineering of Politecnico di Torino, 10129, Torino (Italy). E-mails: \{firstname,lastname\}@polito.it
		}
		\thanks{
			This project has received funding from the European Union’s Horizon Europe research and innovation programme under grant agreement No. 101070238. Views and opinions expressed are however those of the author(s) only and do not necessarily reflect those of the European Union. Neither the European Union nor the granting authority can be held responsible for them.
		}
		\thanks{
			Manuscript received xxx xx, xxxx; revised xxx xx, xxxx.
		}
	}


	\IEEEpubid{0000--0000/00\$00.00~\copyright~2023 IEEE}

	\maketitle

	\section{Introduction}
	\label{sec:intro}

	Integrating \glspl{ann} at the edge is a pivotal computing advancement, enabling direct application of \gls{ai} capabilities in devices and systems at the periphery of networks, resulting crucial in domains like autonomous vehicles\ignore{ \cite{luckow2016deep}}, industrial automation\ignore{ \cite{lu2017industry}}, and healthcare\ignore{ \cite{lu2017industry}}~\cite{chang2021survey}. Using \glspl{ann} at the edge allows systems to process substantial data locally, reducing latency, power consumption, and reliance on external data centers or cloud services. Additionally, it enhances privacy and security by keeping sensitive data within the edge device \ignore{\cite{satyanarayanan2017emergence}}. This approach boosts the effectiveness and agility of embedded applications, paving the way for innovative breakthroughs across sectors and ushering in a new era of intelligent, decentralized computing.

	Neural networks, tailored for diverse computational tasks, include feed-forward networks for pattern recognition, recurrent networks for sequential data, convolutional networks for image analysis, and transformers for \gls{NLP}\ignore{ \cite{gurney2018introduction,10098596}}. Amidst this variety, \glspl{snn} emerge as a promising paradigm for edge computing in resource-constrained environments\ignore{ \cite{10.1007/978-3-030-58607-2_23}}, particularly intriguing for researchers in neurosciences and \gls{ml} \cite{kasabov2019time}. Unlike traditional models, \glspl{snn} closely mimic biological neurons that efficiently process dynamic, time-sensitive data \ignore{ \cite{nunes2022spiking}}, making them valuable in edge computing scenarios, where real-time decision-making is crucial \cite{xue2023edgemap}. 

	In this context, hardware accelerators are vital for \glspl{snn} applications, significantly improving computational efficiency and speed for real-time processing in resource-constrained environments \cite{Dhilleswararao22ab}. However, the flexibility of specialized hardware design poses a challenge, with applications often requiring diverse network architectures, encoding methods, and neuron models. While awaiting the maturity of \gls{snn} accelerators based on emerging technologies, existing literature proposes various digital hardware solutions (refer to \autoref{subsec:related_work})\ignore{  \cite{pavanello_special_2023}}. Unfortunately, these solutions often constrain network topology to circuit architecture, limiting exploration of the broader design space, whether considering \glspl{asic} or \glspl{fpga}\ignore{   \cite{isik2023survey}}. We propose an alternative strategy, optimizing the network architecture for specific needs and leveraging \glspl{fpga} for deploying custom hardware blocks. This approach enables efficient and low-power \gls{snn} inference engines at the edge, supporting real-time data processing. \glspl{fpga} provide high parallelism and reconfigurability, making them ideal for accelerating complex neural network computations with minimal latency. 

	To support this trend, this paper presents \spikerframework{}, a complete framework for generating efficient low-power and low-area customized \gls{snn} accelerators on \glspl{fpga} for inference at the edge. \spikerframework{} introduces several pivotal contributions. At its core, it provides a fully configurable multi-layer hardware architecture implementing both fully connected and recurrent \glspl{snn}. This architecture is a significant step from the preliminary Spiker model, initially presented in \cite{carpegna_spiker_2022}. It introduces a library of highly efficient architectures delving into a range of approximation techniques to implement remarkably low-area and low-power neurons, thus optimizing resource utilization while maintaining high performance.

	Notably, \spikerframework{} brings a complete design framework to the forefront, a comprehensive toolkit for developing complex \glspl{snn} accelerators. This framework empowers researchers and developers to describe target network architectures with great flexibility, enabling the specification of layers, neuron types, and input encoding techniques using a few lines of Python code. Integrating sophisticated off-line server-based training algorithms like \gls{BPTT} \cite{neftci_surrogate_2019} and \gls{stdp} ensures the network has cutting-edge learning capabilities. Additionally, \spikerframework{} emphasizes the significance of optimizing networks through quantization techniques, reducing complexity while judiciously balancing approximation and accuracy. Finally, the framework seamlessly generates a VHDL model of the accelerator, primed for deployment on Xilinx\textsuperscript{\texttrademark} \gls{fpga} boards. These contributions make \spikerframework{} a robust solution in the hardware-accelerated \gls{snn} landscape.
	\spikerframework{} has been tested on the well-known MNIST dataset \cite{lecun_gradient-based_1998} and compared to state-of-the-art \gls{snn} accelerators for \glspl{fpga}, demonstrating superior performance. Moreover, an accelerator for the \gls{SHD} \cite{cramer_heidelberg_2022} has been generated and evaluated to illustrate the framework's flexibility when handling different problems.

	The rest of the paper is organized as follows: \autoref{sec:background} presents some background on \glspl{snn}, focusing on the models used in \spikerframework; \autoref{subsec:related_work} reviews relevant literature on accelerating  \glspl{snn}. Section \ref{sec:spiker_arch} describes the \spikerframework{} architecture, with all the design choices that it involves, and \autoref{sec:configuration_framework} introduces the framework able to configure, design, and generate custom hardware accelerators. Finally, \autoref{sec:experimental_results} presents the results obtained by applying the designed accelerators on the MNIST and the \gls{SHD} and \autoref{sec:conclusions} concludes the paper.


	\section{Background}
	\label{sec:background}

	This section overviews foundational knowledge on \acrshortpl{snn}, required to understand the remaining parts of the paper.

		\subsection{Spiking Neural Networks}
		\label{subsec:snn}

		\glspl{snn} distinguish themselves through their unique information encoding based on \emph{spikes}, inspired by neuroscience. Indeed, this neuron model mimics biological neurons and synaptic communication mechanisms based on action potentials. The information is thus represented as a flow of spikes with various neural coding techniques, shifting the computational complexity from the \emph{spatial} dimension to the \emph{temporal} dimension. Spike encoding methods in \glspl{snn} range from real current or voltage pulses in specialized analog circuits to numerical representations in software or dedicated digital implementations. This paper focuses primarily on the latter. Spikes convey information through temporal organization, and in the digital domain, they can be approximated as binary values: 'one' for received spikes and 'zero' otherwise. 
		Essentially, neurons are translated into compact computational units that exchange data, i.e., their \emph{activations} through binary bit streams.

			\begin{figure}[hbt!]
				\centering
				\begin{subfigure}[c]{0.32\columnwidth}
					\includegraphics[width=\textwidth]{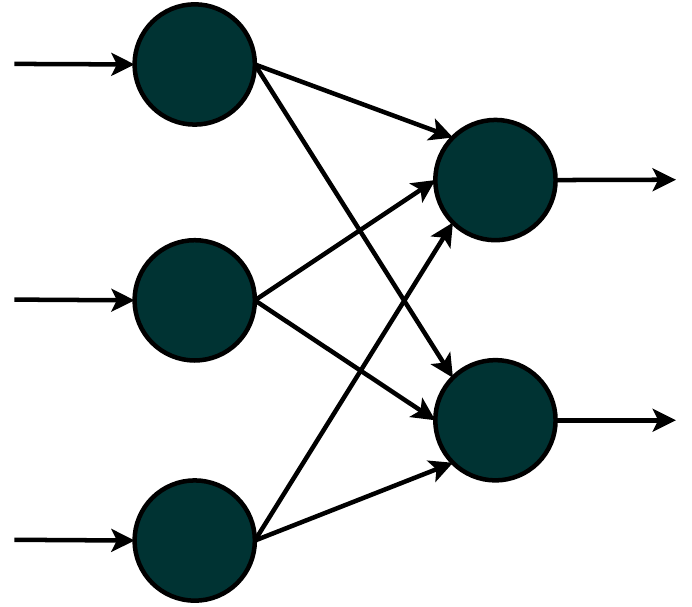}
					\caption{\acrshort{fffc} \gls{snn}}
					\label{fig:fffc_snn}
				\end{subfigure}
				\hfill
				\begin{subfigure}[c]{0.32\columnwidth}
					\includegraphics[width=\textwidth]{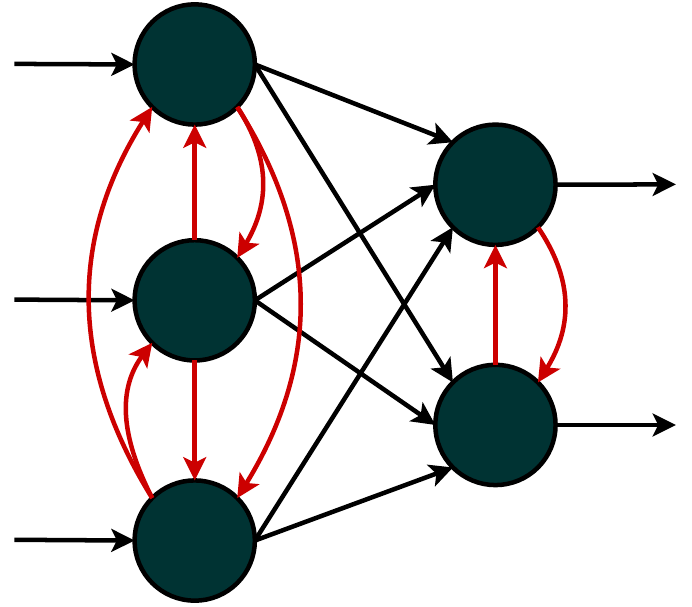}
					\caption{\acrshort{fcr} \gls{snn}}
					\label{fig:rsnn}
				\end{subfigure}
				\hfill
				\begin{subfigure}[c]{0.32\columnwidth}
					\includegraphics[width=\textwidth]{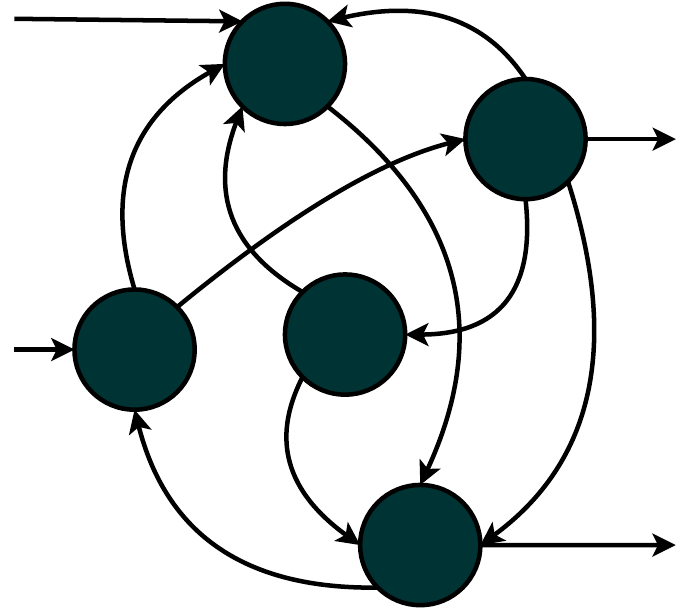}
					\caption{\acrshort{rcr} \gls{snn}}
					\label{fig:random_snn}
				\end{subfigure}

				\caption{Typical architectures of an \gls{snn}}
				\label{fig:snn-archs}
			\end{figure}

		In \glspl{snn}, neuron connections can take various forms, including \gls{fffc} (\autoref{fig:fffc_snn}), \gls{fcr} (\autoref{fig:rsnn}), and \gls{rcr}\ignore{ ensembles (as in reservoir computing, \autoref{fig:random_snn})} organizations\ignore{ \cite{gaurav_spiking_2022}}. \spikerframework{} prioritizes \gls{fffc} and \gls{fcr} structures, ideal for solving diverse classification tasks \cite{yin_accurate_2021}. In particular, in \gls{fcr} \glspl{snn}, neurons connect not only to all nodes in the subsequent layer but also within their layer. This strengthens the joint activity of groups of neurons while inhibiting unrelated ensembles, fostering a competitive environment, which results in a better specialization of the single neurons.


		\subsection{Neuron models}
		\label{subsec:neuron_models}

		Over the past decades, numerous neuron computational models have emerged, originating from nerve electrical conductance measurements and mathematical modeling\ignore{ \cite{torres_modeling_2012}}. The Hodgkin-Huxley model \cite{hodgkinHuxley}, inferred from squid giant axon measurements in 1952, is realistic but complex and has evolved into the simplified Izhikevich model \cite{izhikevich}. However, for hardware implementations, balancing accuracy and complexity commonly leads to using the \gls{lif} model and its simplified version, \gls{if} \ignore{\cite{eshraghian_training_2021}}\cite{brette_adaptive_2005}.
		These models aim to mathematically describe the behavior of a biological neuron focusing on its membrane, which is crucial to defining the internal electrical processes. The membrane selectively permits specific ion passage, accumulating charge and creating a \emph{membrane potential} ($V_m$) defining the neuron's state and behavior.

		The \gls{lif} model in \spikerframework{} encompasses a family of neuron models with varying levels of simplification. \autoref{eq:lif} introduces the discrete-time formulation of a Synaptic Conductance-based II-order \gls{lif} model \cite{brette_adaptive_2005}. Operating in discrete time enables the iterative solution of the differential equations governing the temporal evolution of the membrane potential ($V_m$). In this model, input spikes ($s_{in}$) are integrated by synapses with conductance weights ($W$), influencing the membrane potential based on the input's significance ($W \cdot s_{in}$). The integrated spikes form the synaptic current ($I_{syn}$), which undergoes capacitive discharge ($\alpha \cdot I_{syn}$, with $\alpha < 0$). The membrane integrates this current, resulting in increased or decreased potential based on the current sign. The current sign is influenced by the excitatory or inhibitory nature of the input spikes, impacting the neuron's firing probability. This effect can be positive or negative. The capacitive component ($\beta$) discharges the membrane toward a resting state in the absence of stimuli ($\beta \cdot V_m$, with $\beta < 0$). Lastly, the reset parameter ($r$) models the reset process, as explained later in this section. 
			\begin{equation}
				\begin{cases}
					I_{syn}[n] = \alpha \cdot I_{syn}[n-1] + W \cdot s_{in}[n]  \\
					V_m[n] =(\beta \cdot V_m[n-1] + I_{syn}[n-1]) \cdot r       \\
				\end{cases}
				\label{eq:lif}
			\end{equation}

		An action potential, represented by a separate variable  $s_{out}$, occurs when the membrane voltage ($V_{m}$) surpasses a threshold value ($V_{th}$). This dual-variable approach (membrane and action potential), \ignore{distinguishing between membrane potential and output spike, deviates from the biological modeling where the membrane potential depolarizes along the neuron's axon, generating the action potential. This spike is then propagated to all the neurons connected to the axon, causing the neuron to \emph{fire}. This mathematical model} simplifies the description by treating the spike as a binary variable. \autoref{eq:fire} illustrates the relationship between the output spike and membrane potential $V_m$.

			\begin{equation}
				s_{out}[n] = \begin{cases}
					1, & \mbox{if } V_m > V_{th} \\
					0, & \mbox{if } V_m \leq V_{th}
				\end{cases}
				\label{eq:fire}
			\end{equation}

		Finally, to model the complete discharge of the membrane when the neuron fires, a reset term is used ($r$ in \autoref{eq:lif}). There are different alternatives to applying the reset. \spikerframework{} supports two modes: a \textit{hard-reset}, shown in \autoref{eq:hard_reset}, in which the membrane is instantly brought to zero when a spike is generated, and a \textit{subtractive-reset}, detailed in \autoref{eq:sub_reset}. In the latter, the threshold value is subtracted by the membrane potential.
			\begin{equation}
				r = 1 - s_{out}[n-1]
				\label{eq:hard_reset}
			\end{equation}

			\begin{equation}
				r = 1 - \frac{V_{th}}{\beta \cdot V_m[n-1] + I_{syn}[n-1]}
				\label{eq:sub_reset}
			\end{equation}

		The \gls{lif} model, being a family of models, allows for various simplifications to derive different \gls{lif} descriptions. For instance, setting $\alpha = 0$ transforms the II-order model in \autoref{eq:lif} into a I-order \gls{lif}, where input spikes directly influence the membrane potential. Additionally, with $\beta = 0$, a basic \gls{if} model is obtained, maintaining a constant membrane value without input spikes.

		These different models serve diverse tasks. The II-order \gls{lif} excels in handling input sequences with high temporal information content, capturing longer correlations in precise spike sequences. On the other hand, I-Order \gls{lif}  and \gls{if} models are simpler and preferred for processing static data converted into spike sequences, such as images.
		Working solely with \gls{lif} offers six distinct models: II- and I-order \gls{lif}, and \gls{if}, each with a hard or subtractive reset. These models, supported by \spikerframework{}, combined with the architectures in \autoref{fig:snn-archs}, address various classification and regression problems.

		\subsection{Training}
		\label{subsec:training}

		Training an \gls{snn} involves tuning it for specific problem-solving, such as classifying input data. The training process adjusts synaptic weights ($W$) and internal neuron parameters like threshold ($V_{th}$) and time constants ($\alpha$ and $\beta$) to enhance model accuracy.

		Training \glspl{snn} presents notable challenges, mainly due to the non-differentiability of the \gls{snn} activation function. This paper adopts the Surrogate Gradient approach \cite{neftci_surrogate_2019}, replacing the non-differentiable gradient with a surrogate function, like the spike function itself or a Gaussian function. This enables the application of standard supervised learning techniques, such as \gls{BPTT}, overcoming non-differentiability and facilitating effective \gls{snn} training.

		Alternative solutions, like e-prop \cite{bellec_solution_2020} for \gls{rsnn} and \gls{stdp} for unsupervised weight tuning based on spike timing, exist. However, as \spikerframework{} focuses on inference using pre-trained parameters, these methods fall outside the paper's scope.

	\section{Neuromorphic accelerators: related work}
	\label{subsec:related_work}

\begin{figure}[ht!]
	\centering
	\includegraphics[width=\columnwidth]{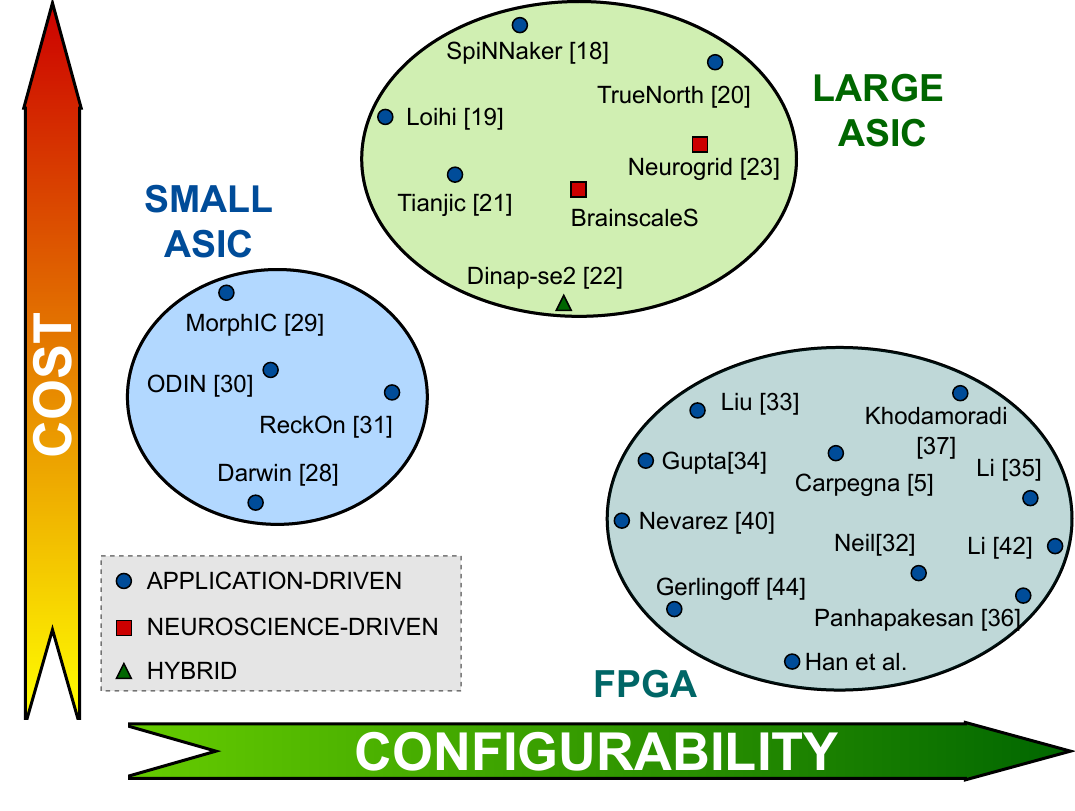}
	\caption{Landscape of neuromorphic hardware}
	\label{fig:sota_overview}
	\end{figure}

	In the past, \glspl{snn} were primarily implemented using software frameworks like Brian/Brian2 \cite{brian2}. However, their unique features, including high parallelism, temporal evolution, and event-driven computation, are ill-suited for dominant Von-Neumann \gls{cpu} architectures with one or a few powerful computational units. Unfortunately, \gls{SIMD} architectures, such as \glspl{GPU} and \glspl{TPU}, optimized for standard \glspl{ann} workloads, are also not well-equipped for efficiently processing event-driven information across multiple timesteps \cite{10.1145/3571155}. Furthermore, the binary spike encoding of \glspl{snn} does not align with the typical 64, 32, or 16-bit numeric representations of these \gls{SIMD} architectures.
	Therefore, dedicated neuromorphic hardware is crucial for ensuring the widespread adoption of \glspl{snn}. \autoref{fig:sota_overview} provides an overview of state-of-the-art accelerators for \glspl{snn}, offering a general perspective rather than an exhaustive list of available solutions. For more detailed information, refer to \cite{basu_spiking_2022}, \ignore{\cite{isik2023survey}, and }\cite{pavanello_special_2023}.

	Contributions in this domain span various design dimensions, including application-driven solutions focused on specific applications and those aimed at modeling biological neuron dynamics. However, this paper primarily emphasizes the hardware technology dimension. The research effort is divided between analog solutions based on emerging technologies and efficient digital implementations \cite{pavanello_special_2023}. In the digital realm, presented solutions differ on the target platform (\gls{asic} or \gls{fpga}) and accelerator size, tailored for either large-scale systems or smal applications.

	Examining large network models, the SpiNNaker system developed at Manchester University is implemented using standard 32-bit ARM M4F \glspl{cpu} simulating neuron activity. It optimizes spike routing between units \ignore{\cite{painkras_spinnaker_2013} }\cite{mayr_spinnaker_2019}.\ignore{The second-generation release in 2019 \cite{mayr_spinnaker_2019} enhances communication between neurons but still operates on a 32-bit ARM M4F architecture that underutilizes the binary nature of spikes.} Promisingly, major computer companies invest in developing their neuromorphic accelerators, such as Intel Loihi \cite{noauthor_next-level_nodate} and IBM True-North \cite{akopyan_truenorth_2015}. These accelerators provide additional optimizations using specialized hardware to execute the neuron model. Another solution, Tianjic \cite{deng_tianjic_2020} from the University of Beijing, aims to implement a hybrid \gls{snn}/\gls{ann} model, benefiting from both domains.

	An alternative design approach is recognizing that real biological neural networks function as analog physical systems. Emulating their efficiency involves using hardware components that approximate biological elements. Chips like Dynap-se2 \cite{richter_dynap-se2_2023} by SynSense, a spinoff of the \gls{ini} in Switzerland, exemplify this concept. Neurogrid \cite{benjamin_neurogrid_2014} from Stanford University and the European project BrainscaleS \cite{pehle_brainscales-2_2022} focus on faithfully simulating portions of a biological brain. Due to their complexity, these systems have programming tools for automatic \gls{snn} configuration. Tools like Rockpool~\cite{noauthor_welcome_nodate} by SynSense and Nengo~\cite{noauthor_nengo_nodate} facilitate automatic configuration and acceleration of \glspl{snn} on neuromorphic hardware based on Python model descriptions.

	While these accelerators suit large-scale neuromorphic systems with good programmability, applications like the \gls{iot}, wearable devices, and biomedical sensors demand smaller sizes and lower power consumption. In such cases, designing specialized hardware accelerators to execute specific tasks, such as classification or regression, efficiently becomes a viable solution.

	To pursue this direction, the first option is designing a compact, programmable \gls{asic} supporting various architectures and models. The focus is on digital solutions for networks ranging from hundreds to a few thousand neurons. A preliminary comparison with non-spiking hardware accelerators, dedicated to efficient convolution execution in \glspl{cnn}, is discussed in \cite{narayanan_spinalflow_2020}, highlighting the energy efficiency of \glspl{snn}. These accelerators often use the \gls{aer} protocol for compatibility with neuromorphic sensors. An example is found in \cite{ma_darwin_2017}, developed at Zhejiang University. Charlotte Frenkel's work at the University of Delft and \gls{eth} Zurich introduces three chips — MorphIC\cite{frenkel_65-nm_2019}, ODIN \cite{frenkel_0086-mm2_2019}, and ReckOn\cite{frenkel_reckon_2022} — exploring online learning on small, low-power, and efficient accelerators.

	The final digital accelerator option, central to this paper, involves developing a specialized \gls{fpga}-based accelerator, offering advantages like cost reduction and increased flexibility by bypassing tape-out design needs. Many \gls{iot} edge systems now integrate \gls{fpga}s for task acceleration, potentially expanding the application of neuromorphic processors. An early example is the event-driven Minitaur \cite{neil_minitaur_2014}, and subsequent alternatives feature diverse update policies, neuron models, architectural choices, and network sizes \cite{liu_low_2023,gupta_fpga_2020,li_firefly_2023,panchapakesan_syncnn_2021,khodamoradi_s2n2_2021,han_hardware_2020,zhang_low-cost_2020,nevarez_accelerating_2021,asgari_low-energy_2020,li_fast_2021,carpegna_spiker_2022}. Another key advantage of using an \gls{fpga} is its intrinsic reconfigurability. It allows hardware reprogramming to modify the network architecture or neuron model, tailoring the accelerator to specific application requirements. However, existing accelerators still need to exploit this characteristic efficiently. The first example of \gls{fpga}-oriented \gls{dse} is found in \cite{abderrahmane_design_2020}, focusing on the best encoding technique for input data translation into spike sequences. Conversely, $E^3NE$\cite{gerlinghoff_e3ne_2022} provides a block library to configure networks for specific applications, optimizing data movement and hardware utilization, with a dedicated section for input encoding.

	In this scenario, there is still a lack of a comprehensive framework that conceals the internal details of the architecture, enabling the user to operate at higher abstraction levels. For instance, a Python description of the model could be automatically translated into custom blocks. \spikerframework{} moves in this direction, trying to address this challenge. 

	\section{\spikerframework{} architecture}
	\label{sec:spiker_arch}

	This section presents the \spikerframework{} hardware architecture, which serves as the central component of the \spikerframework{} \gls{snn} hardware acceleration framework. The architecture is introduced top-down, beginning with the high-level network model and then delving into the neurons and input/output interfaces.

		\subsection{Network architecture}
		\label{subsec:network_architecture}

		The \gls{snn} architecture presented here builds upon the initial Spiker architecture introduced in \cite{carpegna_spiker_2022}. While our earlier work provided a proof of concept tailored for inference on the MNIST dataset \cite{lecun_gradient-based_1998}, derived from the \gls{snn} model by Diehl and Cook \cite{diehl_unsupervised_2015}, \spikerframework{} focuses on a generic and fully configurable architecture adaptable to various problems.

\begin{figure*}[ht!]
				\centering
			
				\includegraphics[width=0.8\textwidth]{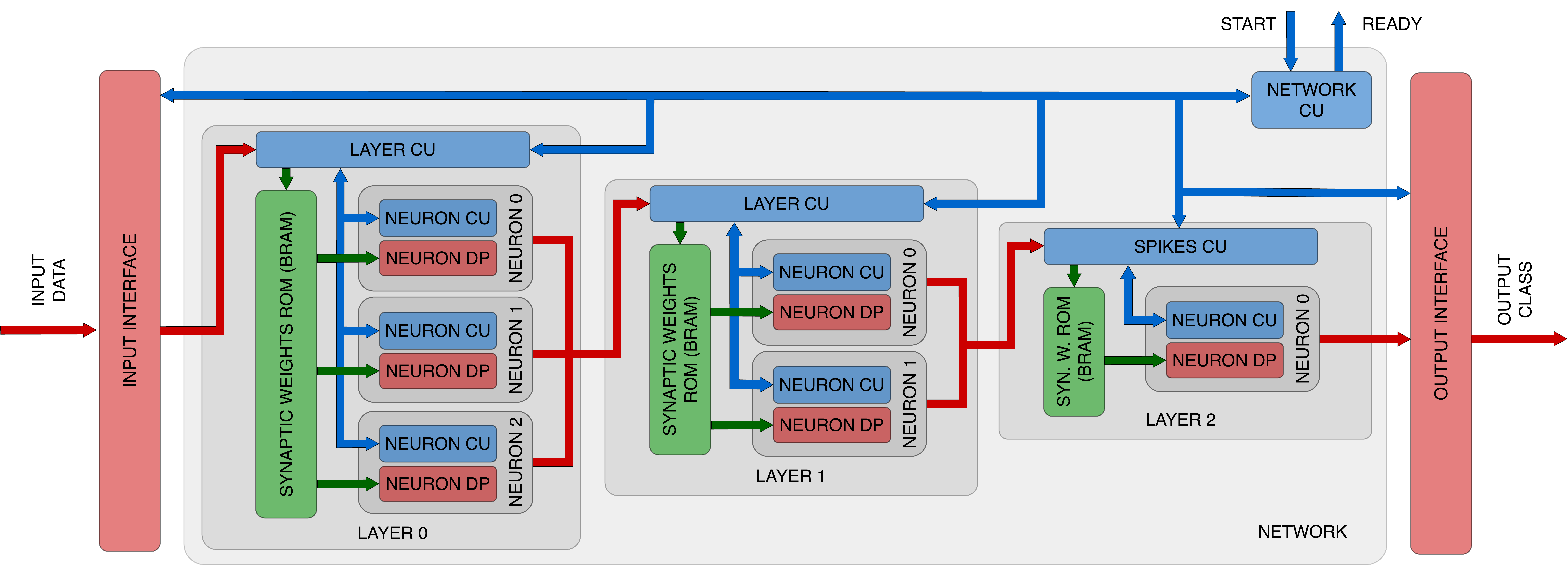}
			
				\caption{\spikerframework{} example of \gls{fffc} architecture. The example includes three layers with different numbers of neurons and depicts all the architecture's main control blocks.}
				\label{fig:full_network}
			\end{figure*}

		\autoref{fig:full_network} depicts the high-level architecture of a toy example of a three-layer \gls{fffc} architecture used to introduce the three hierarchical levels of \glspl{CU} that characterize  \spikerframework{}: (i) the \emph{network \gls{CU}}, responsible for synchronizing the various components within the network; (ii) the \emph{layer \glspl{CU}}, orchestrating the update of the neurons of a layer based on a set of input spikes; (iii) the \emph{neuron \gls{CU}}: the accelerator core controlling the update of the membrane potential in each neuron. This organization represents a highly optimized architecture in terms of performance and space utilization. 

		Block communication is based on a simple two-signal (\texttt{start}/\texttt{ready}) handshake protocol to ensure high modularity while minimizing design complexity. When a block (i.e., a neuron or a layer) is ready to work, it notifies the corresponding \gls{CU} through the \texttt{ready} signal and awaits a new \texttt{start} signal to begin the computation. Consequently, if two blocks need synchronization, combining the two \texttt{ready} signals with an  \texttt{AND} gate ensures that the \gls{CU} waits for both before initiating a new computation. This protocol is also employed at the interface with the external world. Such an approach maintains modularity in the design and paves the way for various architectural solutions.


		\subsection{Network \gls{CU}: global synchronization}
		\label{subsec:central_control}

		The primary function of the Network \gls{CU} is to coordinate the temporal evolution of the neurons of the different layers during an inference. As previously mentioned, in an \gls{snn}, information is encoded as trains of spikes (i.e., sequences of bits) received on every input. Each train is characterized by a given duration (i.e., the number of transmitted bits) denoted as \texttt{N\_cycles}. Thus, the network performs inference by evolving over \texttt{N\_cycles} temporal steps to analyze the temporal patterns. 
			\begin{figure}[ht!]
				\begin{subfigure}{0.4\columnwidth}
					\centering
				
					\includegraphics[width=\columnwidth]{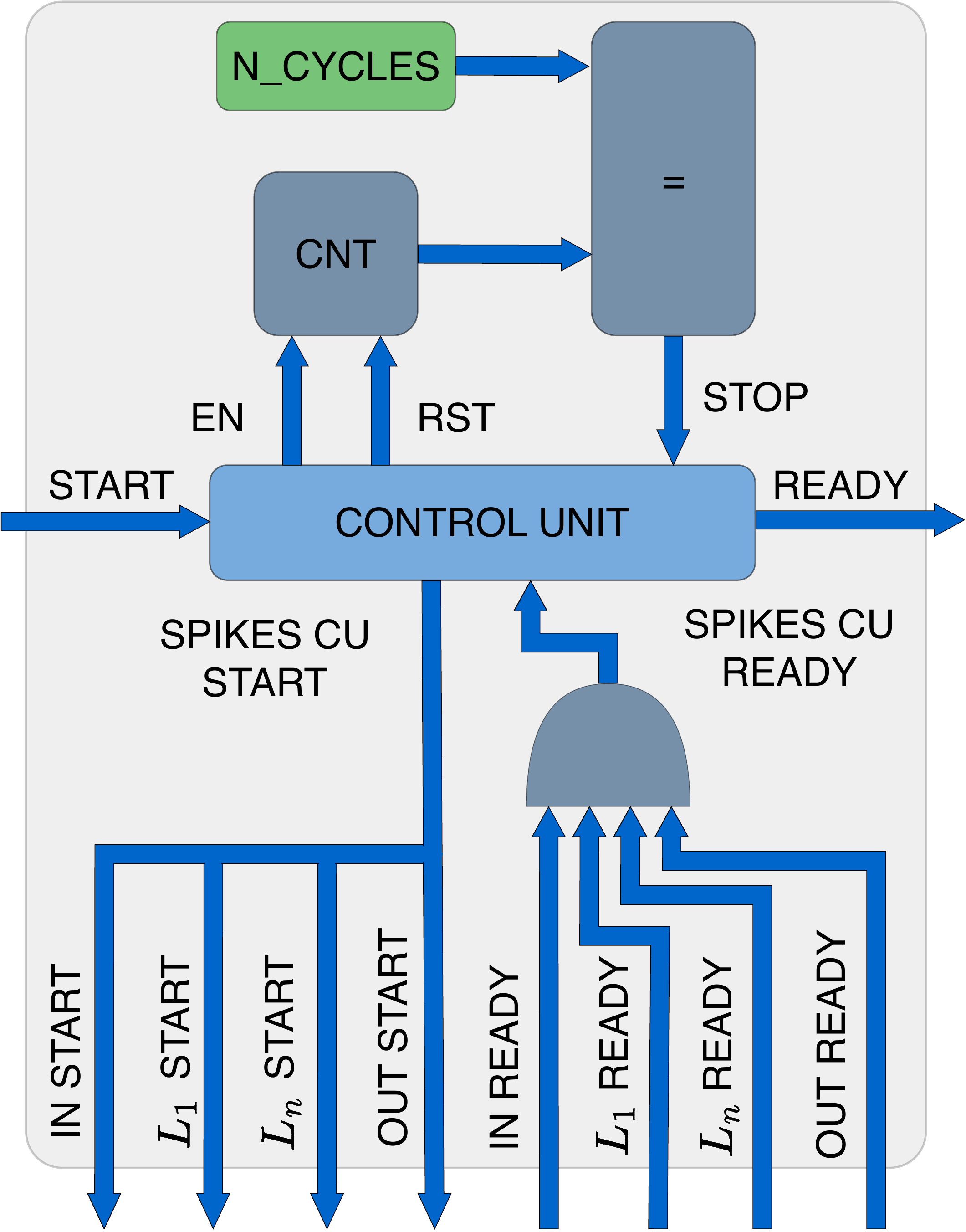}
				
					\caption{Network \gls{CU}}
					\label{fig:central_control}
				\end{subfigure}
				\hfill
				\begin{subfigure}{0.40\columnwidth}
					\centering
				
					\includegraphics[width=\columnwidth]{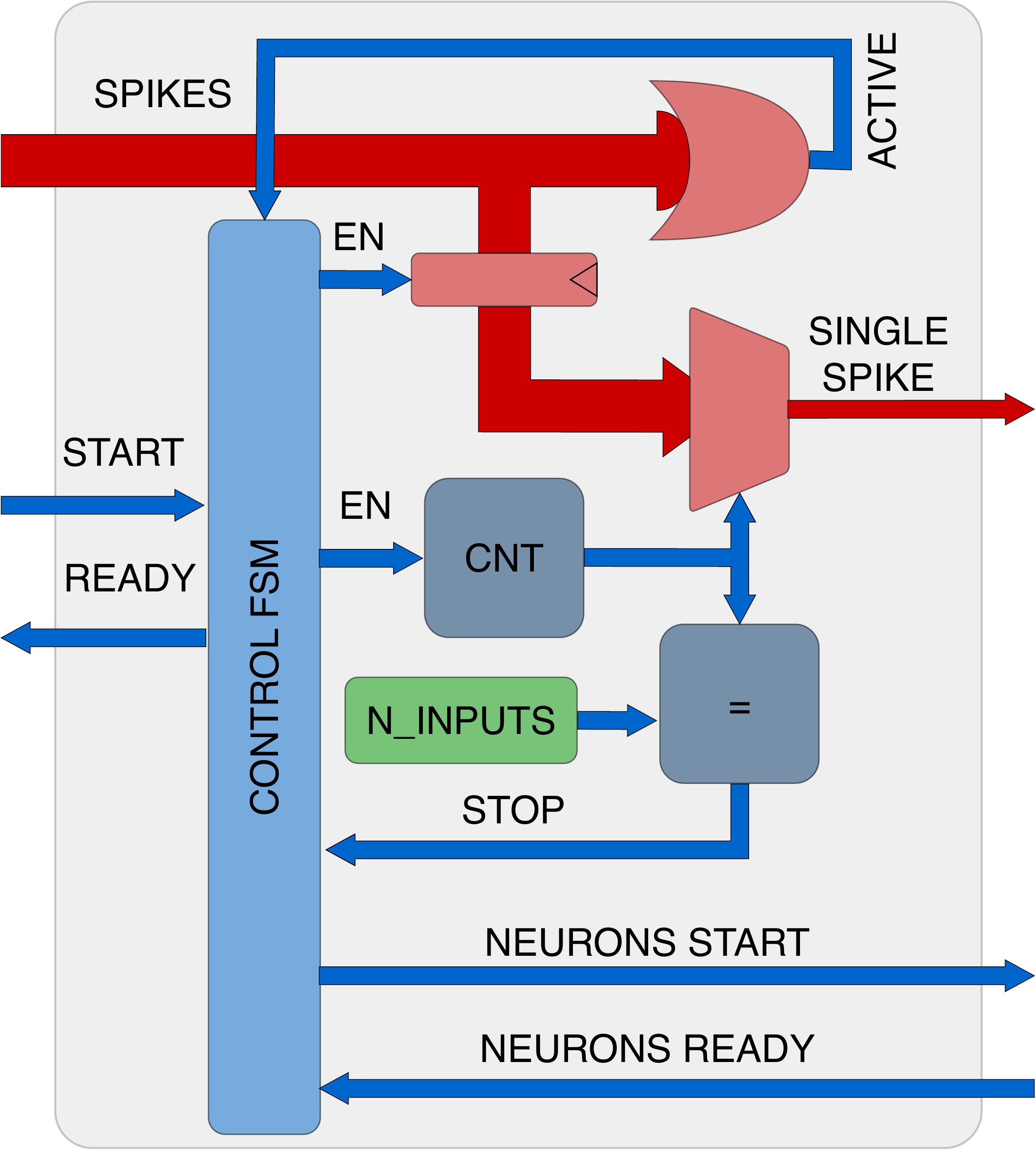}
				
					\caption{Layer \gls{CU}}
					\label{fig:spikes_control}
				\end{subfigure}

				\caption{Internal architecture of the Network and Layer \glspl{CU}}
			\end{figure}
			
		The Network \gls{CU} reported in \autoref{fig:central_control} receives a \texttt{start} signal when a new inference must be initiated. It then manages an iterative process. At every iteration, it awaits the \texttt{ready} signal from all Layer \glspl{CU}  composing the network to ensure all layers are prepared to work (i.e., \texttt{L1 $\dots$ Ln ready}). The Network \gls{CU} also synchronizes with the input/output interface to ensure that data are available and results are correctly delivered (i.e., \texttt{IN/OUT ready}). Subsequently, it asserts a set of \texttt{start} signals, enabling all connected blocks to perform a computation step. At the same time, the internal counter (\texttt{CNT}) increases to track the computation length. Upon reaching the desired cycle count (\texttt{N\_cycles}), the loop concludes, and the \texttt{ready} signal is asserted to indicate the end of the inference process.

		\subsection{Layer \gls{CU}: deliver spikes to neurons}
		\label{subsec:spikes_control}

		In a fully connected multi-layer \gls{snn} as the one proposed in \autoref{fig:full_network}, a parallel update of each layer involves three dimensions: (i) the number of neurons, (ii) the number of inputs processed by each neuron, and (iii) the temporal dimension of each input, representing the number of cycles. 

		The last dimension is inherently sequential and cannot be parallelized, as it depends on the temporal evolution of the inputs. This dimension is managed by the Network \gls{CU} discussed in \autoref{subsec:central_control}, which manages the update during each cycle.

		If the network is sufficiently small, it could be feasible to update all neurons with their inputs in parallel within a single cycle. However, the network and input data sizes are typically too high to achieve such a degree of parallelism. Consequently, \spikerframework{} exploits only one dimension to obtain parallelism, concurrently updating all neurons within a layer while sequentially providing inputs to each neuron. The Network \gls{CU}, depicted in \autoref{fig:spikes_control}, oversees this process.

		Once again, this circuit operates based on a \texttt{start/ready} protocol. The control unit receives a \texttt{start} signal from the Network \gls{CU} and enters a loop: it awaits the readiness of the neurons composing the layer to process a new spike (\texttt{neurons ready} signal), initiates the computation (\texttt{neurons start} signal), and increments the internal counter (\texttt{CNT}). The counter directly selects the spike to be processed from the sampled inputs. When all input spikes have been provided to the neurons (\texttt{N\_INPUTS}), the loop concludes, and the control unit asserts the \texttt{ready} signal.

		An additional component visible in the upper section of \autoref{fig:spikes_control} is an \texttt{OR} gate utilized to verify if there is at least one active spike among the inputs. Currently, no encoding or compression has been applied to the spikes. However, considering that \glspl{snn} typically exhibits sparse activity, avoiding unnecessary computations when there are no spikes can significantly save time and power. The role of the \texttt{OR} gate is to compress along the time dimension: if there is no active spike in a particular cycle, looping over all inputs to provide no spike to the neuron becomes unnecessary.

		\subsection{Neuron models}
		\label{subsec:spiker_neuron_model}

\begin{figure*}[htb!]
				\centering
				\begin{subfigure}[c]{0.20\textwidth}
					\centering

					\includegraphics[width=\textwidth]{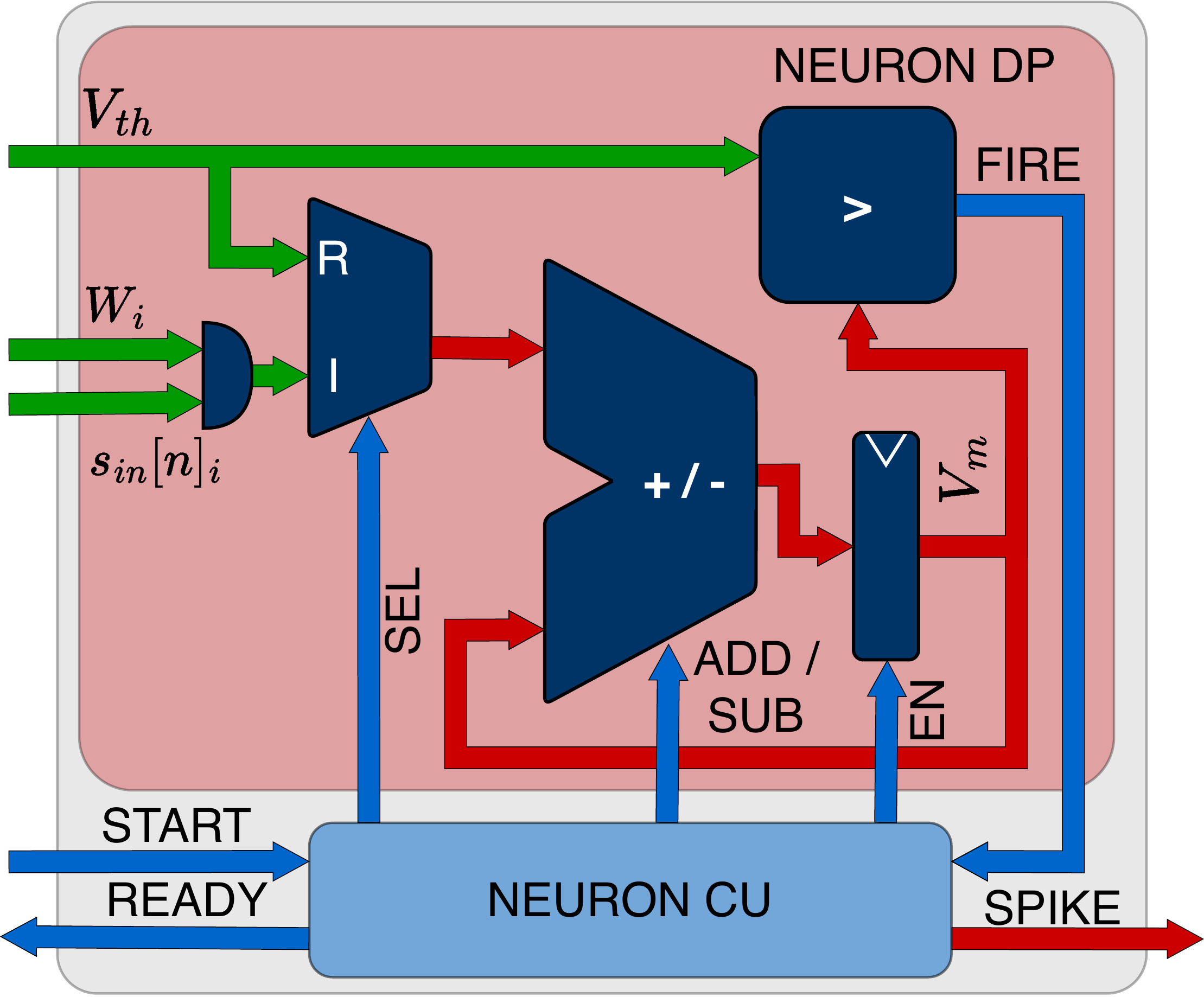}

					\caption{IF neuron with dynamic reset}
					\label{fig:if_rst_dyn}
				\end{subfigure}
				\hfill
				\begin{subfigure}[c]{0.20\textwidth}
					\centering

					\includegraphics[width=\textwidth]{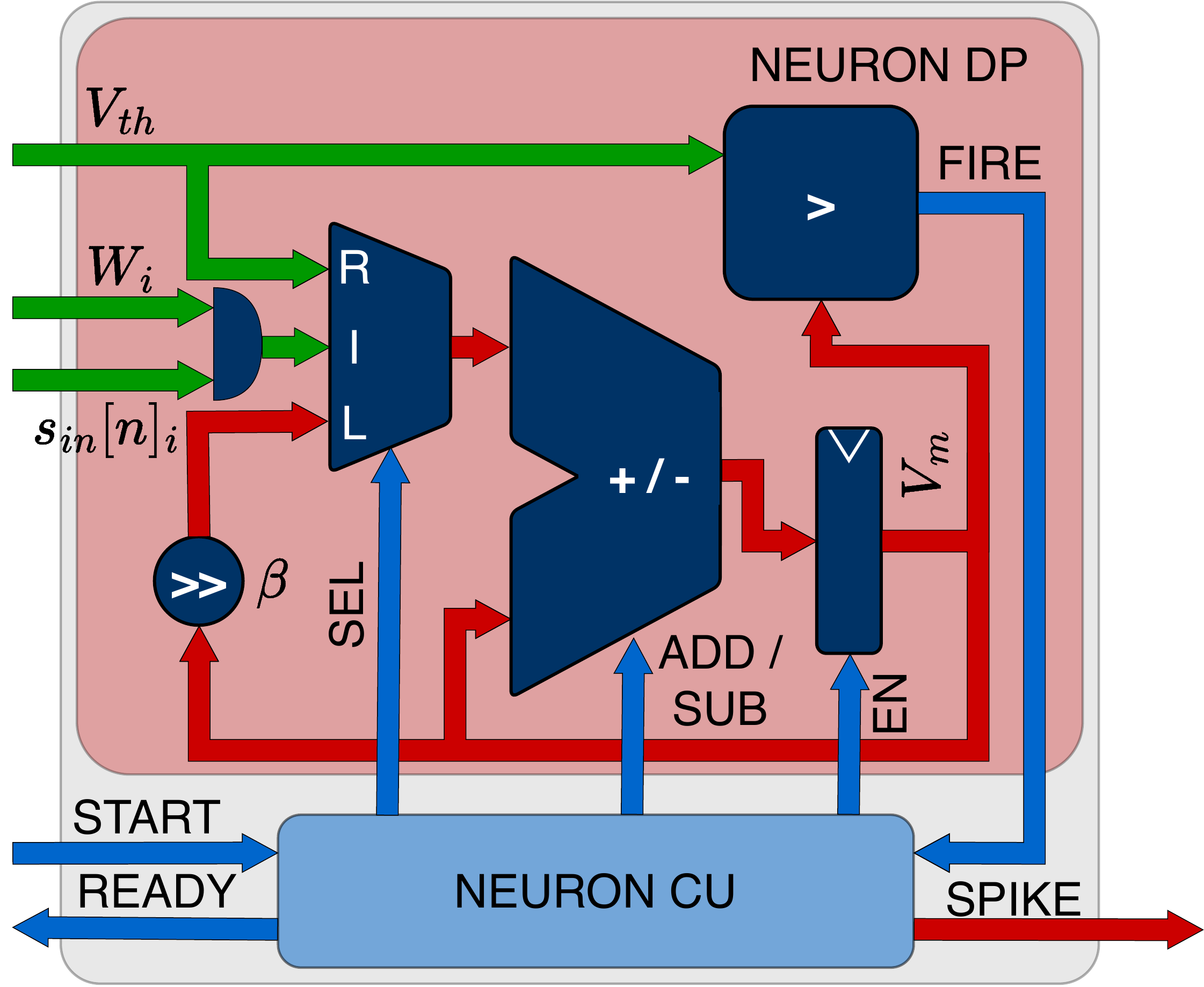}

					\caption{I-order \gls{lif} neuron with dynamic reset}
					\label{fig:lif_ord_I_rst_dyn}
				\end{subfigure}
				\hfill
				\begin{subfigure}[c]{0.40\textwidth}
					\centering

					\includegraphics[width=\textwidth]{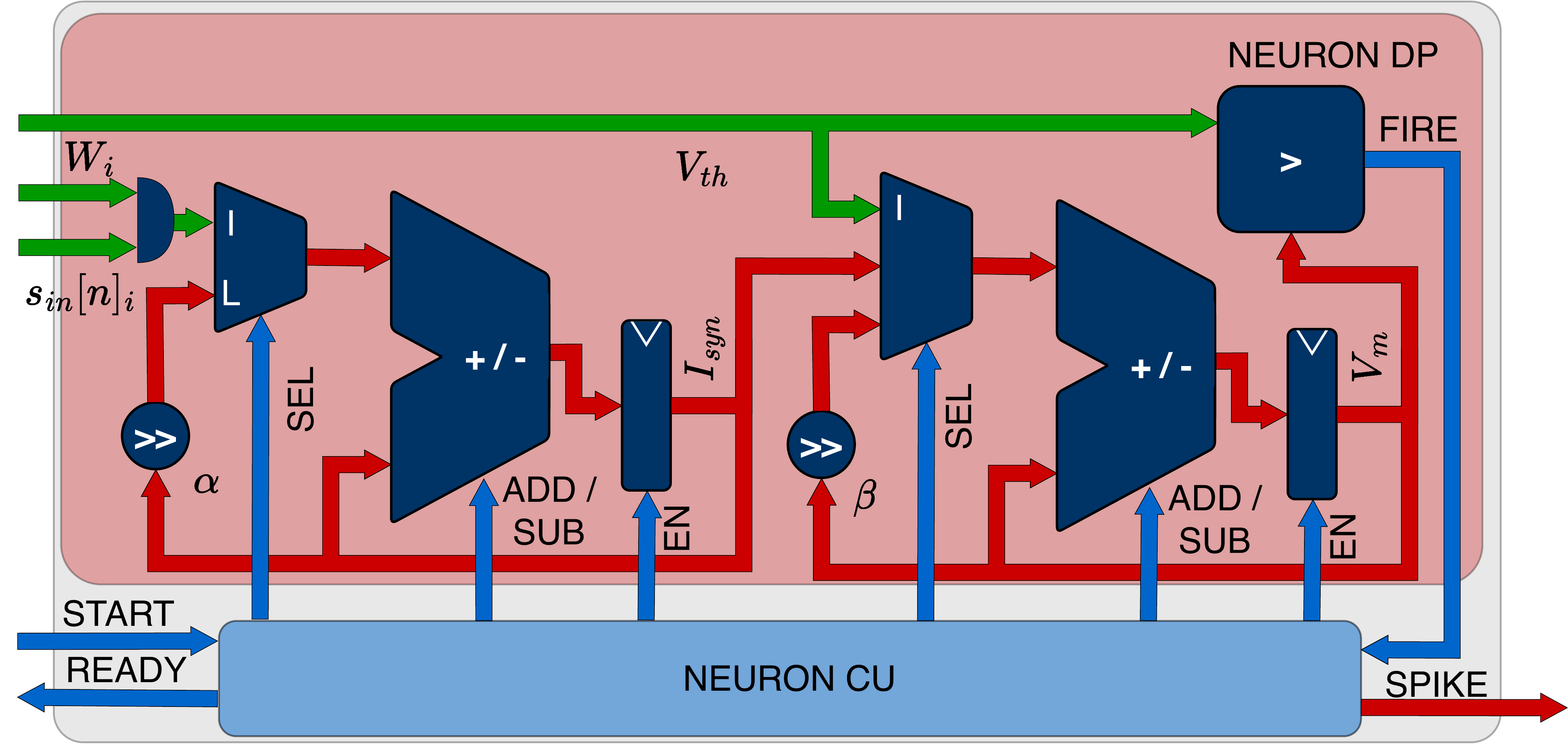}

					\caption{II-order \gls{lif} neuron with dynamic reset}
					\label{fig:lif_ord_II_rst_dyn}
				\end{subfigure}
				\vfill
				\begin{subfigure}[c]{0.20\textwidth}
					\centering

					\includegraphics[width=\textwidth]{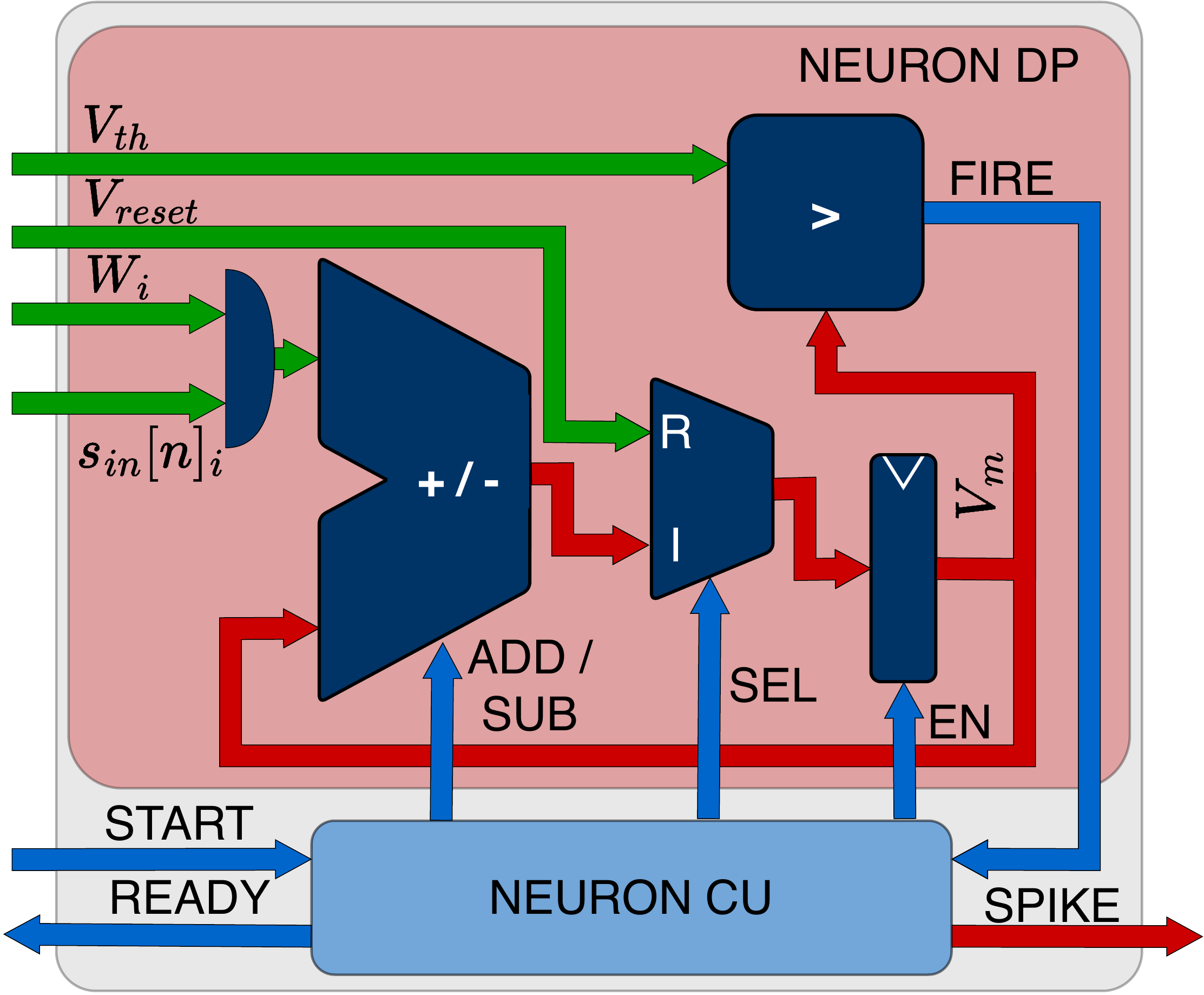}

					\caption{IF neuron with static reset}
					\label{fig:if_rst_fixed}
				\end{subfigure}
				\hfill
				\begin{subfigure}[c]{0.20\textwidth}
					\centering

					\includegraphics[width=\textwidth]{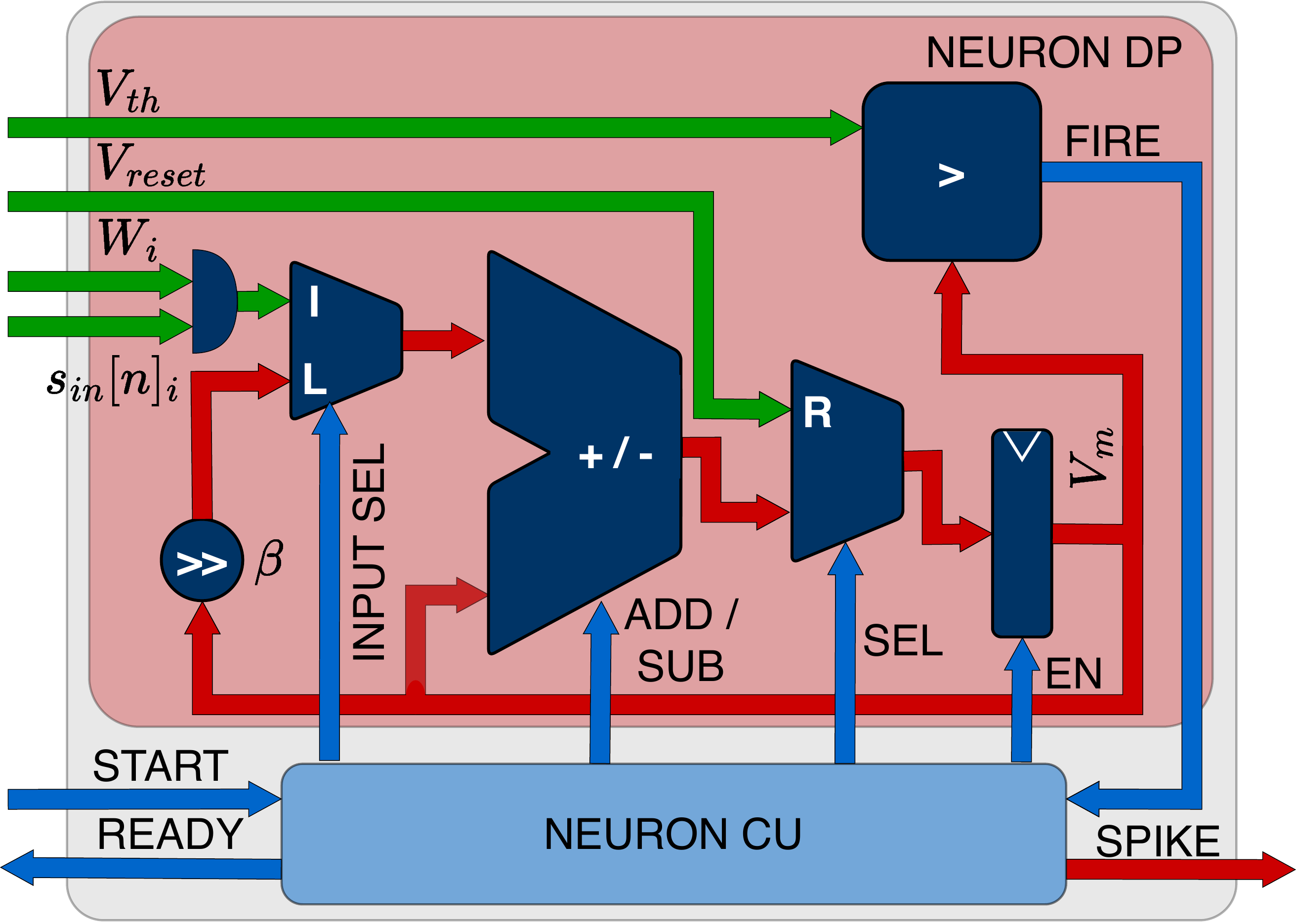}

					\caption{I-order LIF neuron with static reset}
					\label{fig:lif_ord_I_rst_fixed}
				\end{subfigure}
				\hfill
				\begin{subfigure}[c]{0.40\textwidth}
					\centering

					\includegraphics[width=\textwidth]{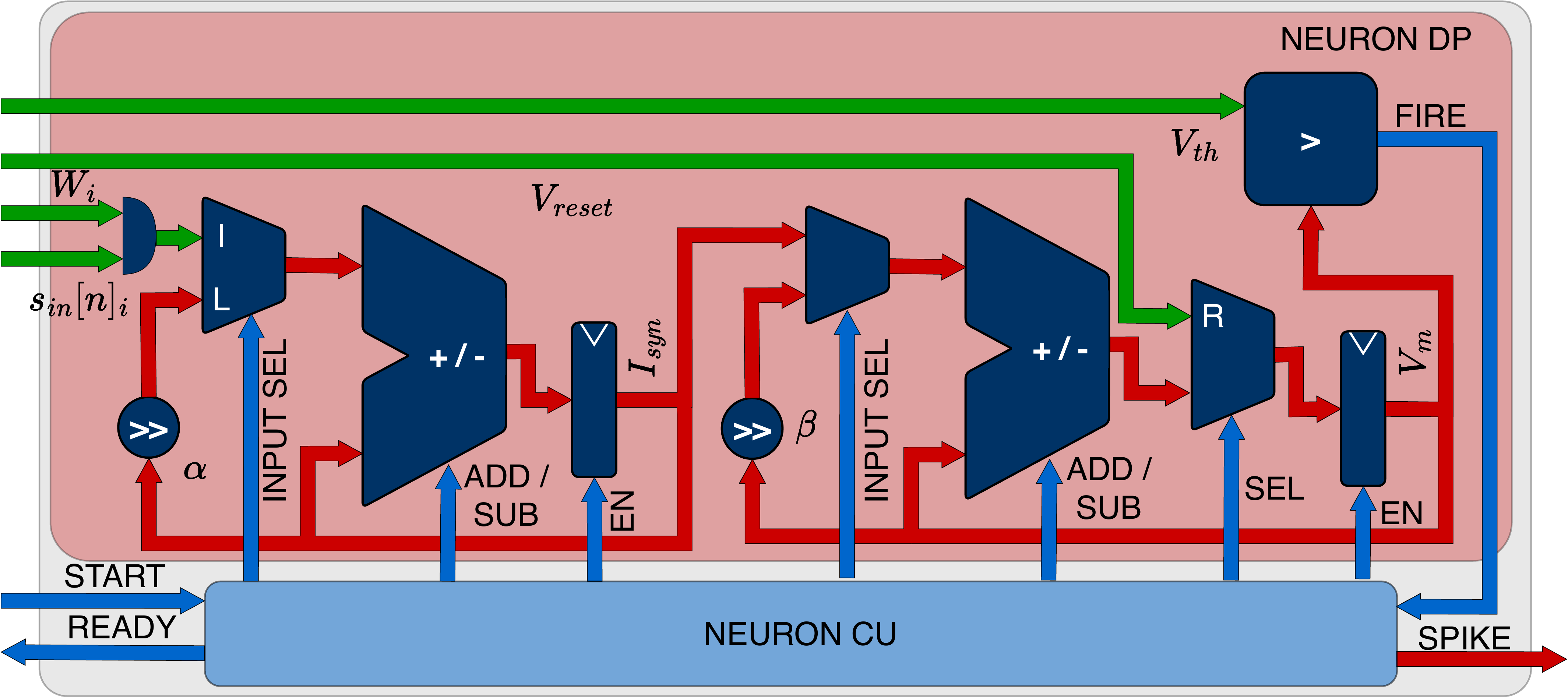}

					\caption{II-order \gls{lif} neuron with static reset}
					\label{fig:lif_ord_II_rst_fixed}
				\end{subfigure}

				\caption{\spikerframework{} neuron architectures in order of increasing complexity from left to right (\gls{if}, I-order \gls{lif} and II-order \gls{lif}), with the subtractive reset on top and the fixed one on the bottom. In the multiplexers, channels labeled with I indicate the beginning of the Integrate path, R the beginning of the Reset path, and L the beginning of the Leakage path.}
				\label{fig:neuron_schematics}
			\end{figure*}

		All the different \gls{lif} neuron models presented in \autoref{subsec:neuron_models} are translated into dedicated hardware implementations in \spikerframework{}, trying to minimize the required components. Building upon the groundwork laid in \cite{carpegna_spiker_2022}, the proposed neuron functions as a \gls{MAC} unit, augmented with additional components and controls to manage its various states. \autoref{fig:neuron_schematics} shows the obtained architectures, in order of increasing complexity from left to right (\gls{if}, I-order \gls{lif} and II-order \gls{lif}), with the subtractive reset on top and the fixed one on the bottom. 

		From a hardware perspective, the most critical factors of the characteristic equation of the neurons are the multiplications. Four of them can be found in \autoref{eq:lif}:

			\begin{enumerate}
				\item The synapses weighting: $W \cdot s_{in}[n]$
				\item The reset: $V[n-1] \cdot r$
				\item The exponential decays: $\alpha \cdot I_{syn}[n-1]$ and \\$\beta \cdot V_m[n-1]$
			\end{enumerate}

		For the first one, exploiting the binary nature of the spikes reduces the operation to a simple selection: zero if there is no spike, $W$ if a spike is present. This can be implemented as a bitwise \texttt{AND} between the weight and the spike.

		The reset operation can be applied in two ways, as shown in equations \ref{eq:hard_reset} and \ref{eq:sub_reset}. The first case exploits the binary nature of the spike: either the membrane is kept at its value or reset to zero, so this is again a selection process more than a multiplication. The hardware implementation is a bit more general since it allows to explicitly choose the value of $V_{reset}$, which in this case can also be different from 0, as shown in figures \ref{fig:if_rst_fixed}, \ref{fig:lif_ord_I_rst_fixed} and \ref{fig:lif_ord_II_rst_fixed}. The second reset method can be obtained by simply subtracting the threshold voltage $V_{th}$ from the computed value of $V_m$, as shown in figures \ref{fig:if_rst_dyn}, \ref{fig:lif_ord_I_rst_dyn} and \ref{fig:lif_ord_II_rst_dyn}.

		At this point, the last critical multiplication is the one required to compute the step-by-step exponential decay of the membrane. The problem exists only for the two \gls{lif} models (in the \gls{if} model, the membrane is kept fixed without stimuli), with one multiplication needed in the I-order version and two multiplications in the II-order one. The criticality is solved once again, exploiting the characteristics of binary operations. If one of the operators is representable as a power of two, the multiplication can be reduced to a simple bit-shift. Since there is no control on the values of $I_{syn}$ and $V_m$, which evolve dynamically during the update of the network, the only parameters on which it is possible to act are the constant hyper-parameters $\alpha$ and $\beta$. The values can vary between 0 and 1, with larger values corresponding to slower exponential decay. Generally, a value near to 1 is observed. In this case $\alpha$ and $\beta$ can be approximated as $\alpha = 1 - \alpha^{\prime}$ and $\beta = 1 - \beta^{\prime}$, where $\alpha^{\prime}$ and $\beta^{\prime}$ are negative powers of 2. As shown in \cite{carpegna_spiker_2022}, the overall accuracy has no notable impact if such an approximation is applied during the training phase.

		\subsection{Synapses}
		\label{subsec:synapses}

		The primary advantage of implementing \glspl{snn} on dedicated hardware, alongside the execution parallelism, lies in the opportunity to integrate memory and computation. On an \glspl{fpga}, this integration can be achieved through two distinct methods.

		For relatively small memory requirements, such as the internal parameters of the neurons, the internal \glspl{LUT} can serve as a viable memory solution. This approach offers superior speed, leveraging \glspl{FF} and registers. However, the available space is limited, primarily due to the necessity of accommodating the logical functions of the network within the \glspl{LUT}.

		In scenarios where a larger memory capacity is necessary, particularly for synaptic weights, many \glspl{fpga} grant access to discrete units of \gls{SRAM} strategically positioned close to the computing elements, commonly referred to as \gls{BRAM}.

		\spikerframework{} provides a synapse interface, implementing the \texttt{start/ready} handshake protocol, and relies on an initialization file containing quantized weights. Weighs are stored into \glspl{BRAM}. \spikerframework{} expects all neurons to access their respective weights in parallel upon activating the ready signal by the synapse; therefore, it strongly relies on the high parallelism provided by on-board \glspl{BRAM}. \spikerframework{} also permits storing weights in an external \glspl{DRAM} when on-board space is insufficient. In such situations, the synaptic interface loads the weights for the current cycle before asserting the ready signal, impacting the accelerator's speed.

		A secondary configurable attribute concerning synapses involves incorporating feedback connections, such as inter-layer inhibitory connections. \spikerframework{} can be configured to include or exclude these connections, depending on the application requirements.

		\subsection{I/O interface}
		\label{subsec:io_interface}

		\spikerframework{} requires an input/output interface to receive data and transmit results. \spikerframework{} supports two scenarios. 
		In the simple scenario, inputs have already been encoded as spikes. For instance, these data may originate from neuromorphic sensors, such as a \gls{DVS} cameras\ignore{ \cite{lichtsteiner_latency_2008}} or a silicon cochlea\ignore{  \cite{liu_asynchronous_2014}}. Alternatively, they could be pre-encoded by an external block before being stored.

		In a more complex scenario, data are stored in a raw numeric format and converted on-board into spike streams. There are different methods available for this conversion, depending on the type of input data\ignore{ \cite{8689349}}\cite{abderrahmane_design_2020}: (i) firing rate coding (i.e., information is encoded using the instantaneous average firing rate), (ii) population rank coding (i.e., information is encoded using the relative firing time of a population of neurons), or (iii) temporal coding (i.e., information is encoded with the exact timing of individual spikes). An efficient rate encoding structure such as the one proposed in \cite{carpegna_spiker_2022} can be directly connected to \spikerframework{}. Furthermore, several possibilities exist concerning data transmission: data may arrive as a continuous stream directly from a sensor or be transmitted through an external link. Alternatively, data may be stored in memory, necessitating memory access by the accelerator.
		To accommodate these diverse scenarios, \spikerframework{} uses the \texttt{start}/\texttt{ready} handshake protocol to manage the communication with the input interface. 

		At the accelerator output, decisions are usually taken based on the firing activity of the last layer. \spikerframework{} implements the output interface using simple counters, one for each output neuron, that can be post-processed outside the accelerator (e.g., the most active neuron wins). This simple implementation is only one possible option, and it can be easily customized to specific needs. The only requirement of the output interface is to implement the \texttt{start}/\texttt{ready} protocol to synchronize with the network.

		It has to be noted that the latency measures reported in the paper have been taken considering input and output interfaces that can keep the pace of the accelerator. Overall, the network performance strongly depends on the interfaces. Results in \autoref{sec:experimental_results} aim to show the maximum performance \spikerframework{} can reach. 

	\section{Configuration framework}
	\label{sec:configuration_framework}

\begin{figure*}[ht!]
			\centering

			\includegraphics[width=0.98\textwidth]{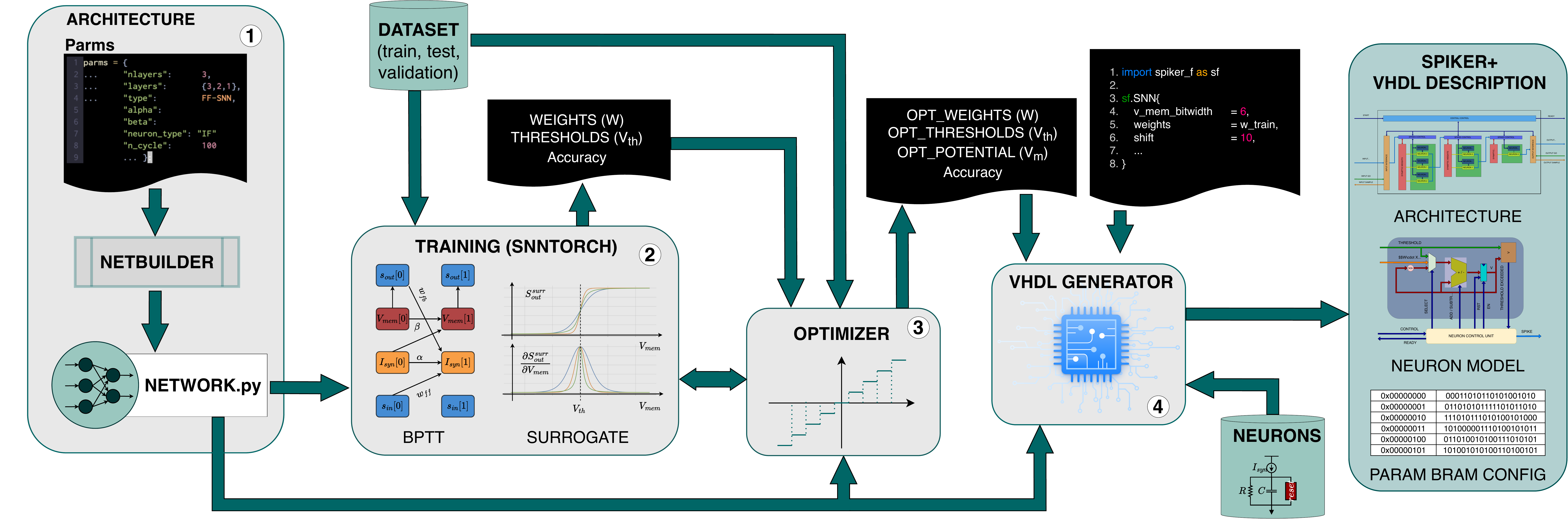}

			\caption{\spikerframework{} configuration framework}
			\label{fig:framework}

		\end{figure*}

	\spikerframework{} goes beyond being a mere hardware accelerator; it is a comprehensive design framework that facilitates easy customization of the \gls{snn} accelerator for specific applications. As detailed in \autoref{sec:spiker_arch}, the platform encompasses six distinct neuron models, a modular layer interface allowing instantiating any desired number of layers, and customizable inter-layer feedback connections. However, manually defining the architecture at the \gls{rtl} requires substantial effort. To tackle this challenge, \spikerframework{} incorporates a Python-based configuration framework, streamlining the customization process to just a few lines of code. The customization and tuning flow for a specific target application within \spikerframework{} is depicted in \autoref{fig:framework}.

	First (step 1), the user selects the target dataset for training and optimizing the network, along with choosing the \gls{snn} model and architecture (e.g., \gls{fffc} or \gls{fcr} \gls{snn}), specifying relevant parameters like the number of layers, neurons per layer, neuron type, and timing constants. These design choices are encapsulated by \texttt{netbuilder} in a Python object serving as a key actor in all subsequent steps. The models in \spikerframework{} align with those in \texttt{snntorch} \footnote{\url{https://snntorch.readthedocs.io/en/latest/}}\ignore{\cite{eshraghian_training_2021}} and similar frameworks.
	Integration with \texttt{snntorch} expedites the development phase, allowing network training using existing algorithms in this open-source Python framework (step 2). However, \spikerframework{} is not limited to this choice; it operates independently of the selected training framework. During the training phase, according to \cite{carpegna_spiker_2022}, time constants $\alpha$ and $\beta$ are first rounded to the closer power of two to compensate during training for this approximation. The result of this phase is the set of trained parameters (e.g., weights and thresholds) and an accuracy assessment of the trained model.

	As training typically occurs in floating-point precision, unsuitable for edge hardware accelerators, \spikerframework{} implements functions to quantize the network, automating the exploration of the quantization impact on \gls{snn} accuracy (step 3). \spikerframework{} uses a two's complement N-bits fixed-point representation. During quantization, if a value exceeds the representable range, it is saturated to the maximum or minimum value, depending on its value. The user can select the desired search interval (default: 64 to 0), and the tool iterates on these values, conducting complete inferences and returning the corresponding accuracy value (step 3). 

	In the final step (step 4), the chosen model, architecture, and trained parameters are delivered to the \texttt{VHDL Generator}, which automatically translates them into a \gls{vhdl} description of the \gls{snn}. This stage capitalizes on the modularity of the proposed architecture and utilizes an available library of neuron models.

	Furthermore, the tool generates configuration files for storing the trained parameters in the \gls{fpga} memory. This provides flexibility to the user in selecting the type of memory. Modern hardware design tools, such as Xilinx Vivado\textsuperscript{\texttrademark}\ignore{ \cite{noauthor_vivado_nodate}}, support the automatic generation of \gls{ROM} memories, allowing users to choose the desired hardware platform, such as \glspl{LUT}, \glspl{BRAM}, or distributed \gls{RAM}. These tools expect a configuration file as input, precisely what \spikerframework{} provides.

	To the best of our knowledge, \spikerframework{} is the first complete high-level synthesis tool for an \gls{snn}, generating \gls{vhdl} code automatically from a high-level Python description of the network. Additionally, the \gls{vhdl} description of the accelerator includes a testbench for enhanced development and simulation convenience. This testbench allows inputs to be read from a file, enabling users to provide the desired input vector stimuli. It provides a straightforward interface to ensure the device synthesizes on the selected \gls{fpga}.

	\section{Experimental results}
	\label{sec:experimental_results}

	\spikerframework{} is evaluated using two widely recognized benchmark datasets: (i) MNIST~\cite{lecun_gradient-based_1998} and (ii) \gls{SHD}~\cite{cramer_heidelberg_2022}.

	MNIST comprises grayscale images of handwritten digits from 0 to 9, commonly used to benchmark \gls{ai} algorithms. This dataset is ideal for comparing \spikerframework{} with other \gls{snn} accelerators. Images are converted into spikes using Poisson-distributed rate encoding. Due to the dataset simplicity, a basic I-order \gls{lif} model with a \gls{fffc} structure suffices for accurate classification.

	\gls{SHD} is explicitly designed as an \gls{snn} benchmark, containing recordings of people pronouncing numbers in English and German. It requires a more complex neuron model, specifically a II-order \gls{lif}, and a network architecture with inter-layer recurrent connections to account for the importance of the time dimension in achieving acceptable classification accuracy.

	These datasets differ significantly, demanding \glspl{snn} with distinct models and complexities, providing an opportunity to test \spikerframework{} reconfigurability for two different tasks. Both models are trained offline employing \gls{BPTT} with two different surrogate gradient functions. Table \ref{tab:exp_setup} summarizes the considered setup.

		\begin{table}[htb]

			\centering

			\caption{Summary of the experimental set-up on the two datastes}
			\label{tab:exp_setup}

				\begin{tabular}{?c?m?s?}

					\hline

					\hline

					\textbf{}				& 
					\textbf{MNIST}				& 
					\textbf{SHD}				\\

					\hline

					\textbf{Type of data}			&
					Grey-scale images			&
					Audio recordings			\\

					\hline

					\textbf{\# inputs}			&
					784					&
					700					\\

					\hline

					\textbf{Spikes time-steps}		&
					100					&
					100					\\

					\hline

					\textbf{Encoding}			&
					Rate code				&
					Custom\cite{cramer_heidelberg_2022}	\\

					\hline

					\textbf{Training samples}		&
					60,000					&
					8,156					\\

					\hline

					\textbf{Test samples}			&
					10,000					&
					2,264					\\

					\hline

					\textbf{Network type}			&
					\gls{fffc} \gls{snn}				&
					\gls{fcr} \gls{snn} \cite{cramer_heidelberg_2022}	
					\cite{yin_accurate_2021}		\\

					\hline

					\textbf{Network arch}			&
					784-128-10				&
					700-200-20				\\

					\hline

					\textbf{Neuron model}			&
					I-order LIF			&
					II-order LIF			\\

					\hline

					\textbf{Reset}				&
					Subtractive				&
					Subtractive				\\

					\hline

					\textbf{Training method}		&
					BPTT with SG				&
					BPTT with SG				\\

					\hline

					\textbf{Surrogate function}		&
					Arctan					&
					Fast Sigmoid				\\

					\hline

					\textbf{Model accuracy}			&
					96.83\%					&
					75.44\%					\\

					\hline

					\hline

				\end{tabular}

		\end{table}

	The remainder of this section is structured as follows: \autoref{subsec:benchmarking} presents results from tuning \spikerframework{} on the two target datasets and compares it with state-of-the-art \gls{snn} accelerators on \gls{fpga}. Subsections \ref{subsec:activity}, \ref{subsec:quantization}, and \ref{subsec:optimal_sizing} analyze various \spikerframework{} configurations, examining the influence of architectural choices and data characteristics on area, latency, and power consumption.

		\subsection{Benchmarking}
		\label{subsec:benchmarking}

		\autoref{tab:MNISTcomparison} provides a comprehensive comparison of \spikerframework{} with recent state-of-the-art \gls{fpga} accelerators designed for \glspl{snn} on the MNIST dataset. The table is split into two sections. The upper section covers \gls{scnn} accelerators, where spiking layers are strategically placed after or interleaved with standard convolutional layers, gradually identifying key features in input images. The lower section considers pure spiking accelerators with fully connected layers of either \gls{if} or \gls{lif} neurons. Notably, the comparison focuses on works published from 2020 onward, while references such as \cite{carpegna_spiker_2022} and \cite{li_fast_2021}\ignore{, and \cite{isik2023survey}} provide information on the performance of older accelerators.

			\begin{table*}[htb]

				\caption{Comparison of \spikerframework{} to state-of-the-art \gls{fpga} accelerators for \glspl{snn}}
				\label{tab:MNISTcomparison}

				\begin{adjustbox}{width=\textwidth, center}

					\begin{tabular}{?m?c|c|c|c|c|c?}

						\hline

						\hline

				\rowcolor{mnist_blue!40}

						\textbf{Design}				& 
						Liu et al.\cite{liu_low_2023}		&
						Nevarez et al.\cite{nevarez_accelerating_2021}	&
						Li et al.\cite{li_firefly_2023}		&
						Gerlinghoff et al.\cite{gerlinghoff_e3ne_2022}		&
						Panchapakesan et al.\cite{panchapakesan_syncnn_2021}				&
						Khodamoradi et al.\cite{khodamoradi_s2n2_2021}				\\

						\hline

						\hline

						\textbf{Year}				& 
						2023					&
						2021					&
						2023					&
						2022					&
						2021					&
						2021					\\

						\hline

						\textbf{$f_{clk}$[MHz]}			& 
						100					&
						200					&
						300					&
						200					&
						200					&
						N/R					\\

						\hline

						\textbf{Neuron bw}			&
						8					&
						8					&
						12					&
						N/R					&
						4					&
						N/R					\\

						\hline

						\textbf{Weights bw}			&
						8					&
						8					&
						8					&
						3					&
						4					&
						N/R					\\

						\hline

						\textbf{Update}				&
						Clock					&
						Clock					&
						Clock					&
						Clock					&
						Event					&
						Event					\\

						\hline

						\textbf{Model}				&
						IF					&
						\gls{sbs}				&
						LIF					&
						LIF\ignore{\cite{wang_efficient_2022}}		&
						IF					&
						LIF					\\

						\hline

						\textbf{FPGA}				&
						XA7Z020					&
						XC7Z020					&
						XCZU3EG					&
						XCVU13P					&
						XCZU9EG					&
						XA7Z020					\\

						\hline

						\textbf{Avail. BRAM}			&
						140					&
						140					&
						216					&
						2688					&
						912					&
						140					\\

						\hline

						\textbf{Used BRAM}			&
						N/R					&
						16					&
						50					&
						N/R					&
						N/R					&
						40.5					\\

						\hline

						\textbf{Avail. DSP}			&
						220					&
						220					&
						360					&
						12288					&
						2520					&
						220					\\

						\hline

						\textbf{Used DSP}			&
						0					&
						46					&
						288					&
						0					&
						N/R					&
						11					\\

						\hline

						\textbf{Avail. logic cells}		&
						159,600					&
						159,600					&
						211,680					&
						3,088,800				&
						822,240					&
						159,600					\\

						\hline

						\textbf{Used logic cells}		&
						27,551					&
						23704					&
						15,000					&
						51,000					&
						N/R					&
						39,368					\\

						\hline

						\textbf{Arch}				&
						\begin{tabular}{c}
							28×28-32c3-p2-\\
							32c3-p2-256-10
						\end{tabular}				&

						\begin{tabular}{c}
							28x28x2-32c5-p2-\\
							64c5-p2-1024-10
						\end{tabular}				&

						\begin{tabular}{c}
							28x28-16c3-64c3-\\
							p2-128c3-p2-\\
							256c3-256c3-10
						\end{tabular}				&
						
						\begin{tabular}{c}
							32×32×1–6c5–p2–\\
							16c5–p2–120c5–\\
							120–84–10
						\end{tabular}				&
					
						\begin{tabular}{c}
							28x28-32c3-p2-\\
							32c3-p2-256-10
						\end{tabular}				&

						\begin{tabular}{c}
							28x28-16c7-24c7-\\
							32c7-10
						\end{tabular}				\\


						\hline

						\textbf{\#syn}				&
						8,960                		&
						75,776		&
						2,560		&
						25,320		&
						10,752		&
						320		\\

						\hline

						\textbf{$T_{lat}$/img [ms]}		&
						0.27					&
						1.67					&
						0.49					&
						0.29					&
						0.08					&
						N/R					\\

						\hline

						\textbf{Power [W]}			&
						0.28					&
						0.22					&
						2.55					&
						3.40					&
						N/R					&
						N/R					\\

						\hline

						\textbf{E/img [$mJ$]}			&
						0.076					&
						0.37					&
						1.250					&
						0.986					&
						N/R					&
						N/R					\\

						\hline

						\textbf{E/syn [$nJ$]}&
						8.48        		&
						4.88        		&
						488         		&
						38.9		&
						N/R					&
						N/R					\\

						\hline



						\textbf{Accuracy}			&
						99.00\%					&
						98.84\%					&
						98.12\%					&
						99.10\%					&
						99.30\%					&
						98.50\%					\\

						\hline

						\hline

				\rowcolor{mnist_blue!40}

						\textbf{Design}				& 
						Han et al.\cite{han_hardware_2020}	&
						Gupta et al. \cite{gupta_fpga_2020}	&
						Li et al.\cite{li_fast_2021}		&
						Carpegna et al.\cite{carpegna_spiker_2022}				&
						\multicolumn{2}{|c|}{\textbf{SPIKER+ (this work)}}		\\

						\hline

						\hline

						\textbf{Year}				& 
						2020					&
						2020					&
						2021					&
						2022					&
						\multicolumn{2}{|c|}{2024}		\\

						\hline

						\textbf{$f_{clk}$[MHz]}			& 
						200					&
						100					&
						100					&
						100					&
						\multicolumn{2}{|c|}{100}		\\

						\hline

						\textbf{Neuron bw}			&
						16					&
						24					&
						16					&
						16					&
						\multicolumn{2}{|c|}{6}			\\

						\hline

						\textbf{Weights bw}			&
						16					&
						24					&
						16					&
						16					&
						\multicolumn{2}{|c|}{4}			\\

						\hline

						\textbf{Update}				&
						Event					&
						Event					&
						Hybrid					&
						Clock					&
						\multicolumn{2}{|c|}{Clock}		\\

						\hline

						\textbf{Model}				&
						LIF					&
						LIF\ignore{\cite{iakymchuk_simplified_2015}} 	&
						LIF					&
						LIF					&
						\multicolumn{2}{|c|}{LIF}		\\

						\hline
						\textbf{FPGA}				&
						XC7Z045					&
						XC6VLX240T				&
						XC7VX485			&
						XC7Z020				&
						\multicolumn{2}{|c|}{XC7Z020}		\\

						\hline

						\textbf{Avail. BRAM}			&
						545					&
						416					&
						2,060					&
						140					&
						\multicolumn{2}{|c|}{140}		\\

						\hline

						\textbf{Used BRAM}			&
						40.5					&
						162					&
						N/R					&
						45					&
						\multicolumn{2}{|c|}{\textbf{18}}	\\

						\hline

						\textbf{Avail. DSP}			&
						900					&
						768					&
						2,800					&
						220					&
						\multicolumn{2}{|c|}{220}		\\

						\hline

						\textbf{Used DSP}			&
						0					&
						64					&
						N/R					&
						0					&
						\multicolumn{2}{|c|}{0}	\\

						\hline

						\textbf{Avail. logic cells}		&
						655,800					&
						452,160					&
						485,760					&
						159,600					&
						\multicolumn{2}{|c|}{159,600}		\\

						\hline

						\textbf{Used logic cells}		&
						12,690					&
						79,468					&
						N/R					&
						55, 998					&
						\multicolumn{2}{|c|}{\textbf{7,612}}	\\

						\hline

						\textbf{Arch}				&
						784-1024-1024-10			&
						784-16					&
						784-200-100-10				&
						784-400					&
						\multicolumn{2}{|c|}{784-128-10}	\\

						\hline

						\textbf{\#syn}				&
						1,861,632				&
						12,544					&
						177,800					&
						313,600					&
						\multicolumn{2}{|c|}{101,632}		\\

						\hline

						\textbf{$T_{lat}$/img [ms]}		&
						6.21					&
						0.50					&
						3.15					&
						0.22					&
						\multicolumn{2}{|c|}{0.78}		\\

						\hline

						\textbf{Power [W]}			&
						0.477					&
						N/R					&
						1.6					&
						59.09					&
						\multicolumn{2}{|c|}{\textbf{0.18}}		\\

						\hline

						\textbf{E/img [$mJ$]}			&
						2.96					&
						N/R					&
						5.04					&
						13					&
						\multicolumn{2}{|c|}{0.14}		\\

						\hline

						\textbf{E/syn [$nJ$]}&
						1.59					&
						N/R					&
						28					&
						41					&
						\multicolumn{2}{|c|}{1.37}		\\

						\hline



						\textbf{Accuracy}			&
						97.06\%					&
						N/R					&
						92.93\%					&
						73.96\%					&
						\multicolumn{2}{|c|}{93.85\%}		\\

						\hline
						
						\hline

					\end{tabular}

				\end{adjustbox}
			\end{table*}

		In image classification tasks, \glspl{scnn} accelerators achieve superior accuracy, peaking at 99.30\% with larger and more complex networks, often utilizing DSP cores on the \gls{fpga} (e.g., \cite{nevarez_accelerating_2021, khodamoradi_s2n2_2021, li_firefly_2023}). However, despite \spikerframework{} being explicitly designed to trade off accuracy with other design dimensions (i.e., power, area, latency), it still achieves a commendable accuracy of 93.85\%. Among purely spiking accelerators, it is only surpassed by \cite{han_hardware_2020}, which employs an accelerator with higher latency, power, and size. Remarkably, \spikerframework{} excels in compactness and low power consumption. On a low-end Xilinx\textsuperscript{\texttrademark} XC7Z020 \gls{fpga} board, it uses only 7,612 logic cells (4.8\% of available cells) and 18 \glspl{BRAM} (13\% of available \glspl{BRAM}), with a total power requirement of 180mW. This makes it an optimal solution for limited space or power-constrained applications. It is important to note that \autoref{tab:MNISTcomparison} measures the area with a general "logic cells" value obtained by combining \glspl{LUT} and \glspl{FF} for a concise overview. For \spikerframework{}, these values are 4,314 and 3,298, respectively. Detailed values for other accelerators can be found in their respective papers.

		A noteworthy observation from the comparison in \autoref{tab:MNISTcomparison} is that the top power-efficient accelerators employ a clock-driven update policy. This counterintuitive finding contradicts the general literature assertion favoring event-driven approaches for power efficiency. A clock-driven approach seems the most effective solution for relatively small-sized accelerators lacking sufficient sparsity in spiking activities. Li et al. attempted to address this with a hybrid architecture, adapting its update strategy to input sparsity. However, this strategy did not yield significant power savings, even with added complexity \cite{li_fast_2021}. Importantly, the power consumption of the accelerator reported in \cite{carpegna_spiker_2022} is unrealistically high. This is due an error in mapping I/O ports during the accelerator implementation, leading to an overestimated value.

		Regarding latency, \spikerframework{} requires $780\mu s$ to classify an input image. Although not the fastest result in \autoref{tab:MNISTcomparison}, this achievement is noteworthy considering the limited hardware resources and power consumption. Factors influencing this latency include: (i) the clock frequency capped at 100MHz due to \gls{BRAM} access time; (ii) the image encoding using 100 time-steps (the window size impacts inference time directly); (iii) the speed of the input and output interfaces; (iv) the input spiking activity affecting the classification time, as explained in \autoref{subsec:activity}. For comparison, Carpegna et al. \cite{carpegna_spiker_2022} achieved $220\mu s$ image classification time with a 3500 time-step encoding window. This work employed a different encoding that privileged spike sparsity, impacting accuracy (i.e., 73.96\%) but demonstrating how sparsity can reduce inference time.

		In the second use case, \spikerframework{} is the first accelerator tested on \gls{SHD}. Consequently, results reported in \autoref{tab:SHDbenchmark} are presented independently, without comparison to other architectures. 

			\begin{table}[htb]

				\centering

				\caption{Benchmarking on the \gls{SHD} dataset}
				\label{tab:SHDbenchmark}

				\begin{tabular}{?s?c?s?c?}

					\hline

					\hline


					\hline

					\textbf{$f_{clk}$[MHz]}			& 
					100					             &
			    \textbf{Avail. logic cells}		&
					159,600					\\

					\hline

					\textbf{Neuron bw}			    &
					8					           &
			    \textbf{Used logic cells}		&
					18,268					\\

					\hline

					\textbf{FF weights bw}			&
					6					              &
			    \textbf{Arch}				&
					700-200-20				\\

					\hline
					
					\textbf{RR weights bw}			&
					5					           &
			    \textbf{\#syn}				&
					184,000					\\

					\hline

					\textbf{Update}				&
					Clock					   &
			    \textbf{$T_{lat}$/img [ms]}		&
					0.54					\\

					\hline

					\textbf{Model}				&
					II order LIF				&
			    \textbf{Power [W]}			&
					0.43					\\

					\hline

					\textbf{FPGA}				&
					XA7Z020					     &
			    \textbf{E/img [$mJ$]}			&
					0.23					\\

					\hline

					\textbf{Avail. BRAM}			&
					140					         &
			    \textbf{E/syn [$nJ$]}			&
					1.25                 		\\

					\hline

					\textbf{Used BRAM}			&
					51					          &
			    \textbf{Accuracy}			&
					72.99\%					\\

					\hline


					\hline

					\hline

				\end{tabular}

			\end{table}

		The hardware requirements for this model exceed those used for MNIST due to several factors: the network architecture required to process this complex dataset is larger, the neuron model is a more complex II-order \gls{lif} (i.e., double size compared to a I-order \gls{lif}), and the accelerator uses a \gls{fcr} model featuring inter-layer feedback connections with weights stored in \gls{BRAM}. The bit widths of the neuron membrane potential and weights are also higher. However, \spikerframework{} remains smaller and more power efficient than most accelerators in \autoref{tab:MNISTcomparison}. Latency is reduced from $780\mu s$ for MNIST to $540\mu s$ for \gls{SHD} due to lower input activity in the biologically inspired encoding used for this dataset.

		\subsection{Performance vs input activity}
		\label{subsec:activity}

\begin{figure}[ht!]
				\centering
				
				\includegraphics[width=0.98\columnwidth]{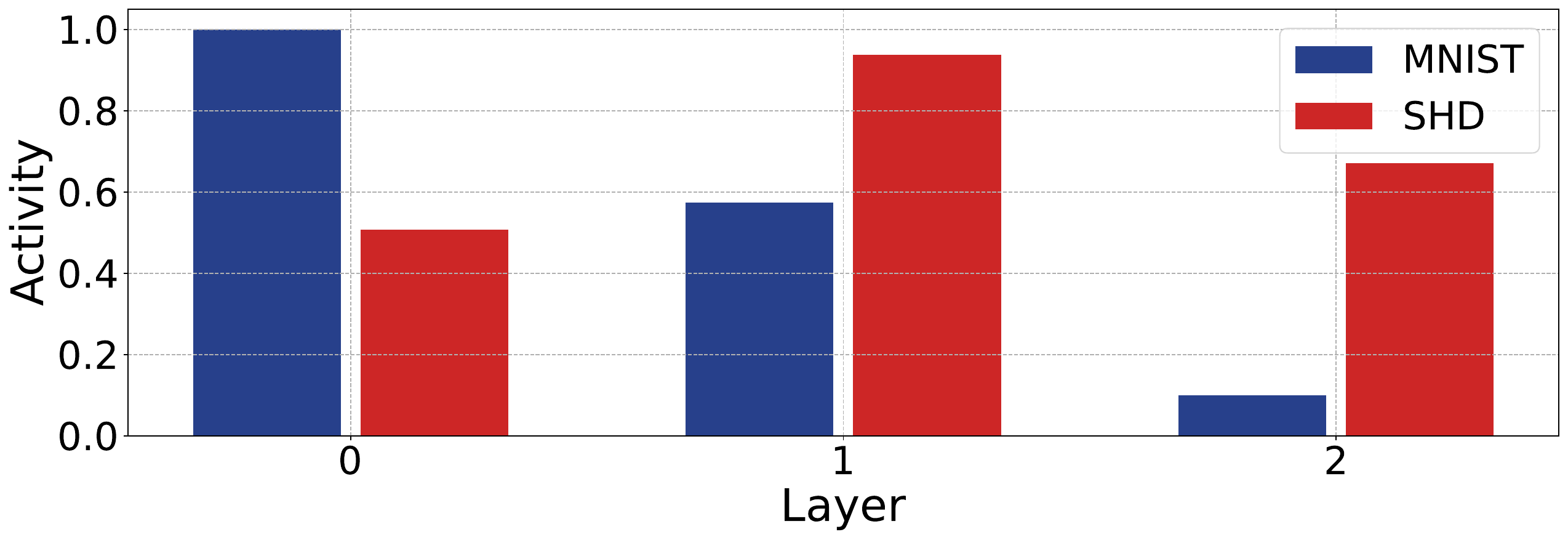}

				\caption{Visual representation of the average number of active cycles at various stages of the network}
				\label{fig:activity}
			\end{figure}

		As introduced earlier, the input spiking activity influenced by the encoding method significantly impacts the accelerator's performance. Before diving into this analysis, \autoref{fig:activity} visually represents the average number of active cycles at different network layers. A notable difference is observed between the two considered datasets. In MNIST, the activity decreases monotonically across the network, while for \gls{SHD}, there is a peak of activity in the first hidden layer. This difference may arise from the inter-layer feedbacks in \gls{SHD}, leading to higher joint activity than the \gls{fffc} architecture used in MNIST.

		Since all layers update in parallel and process inputs sequentially, latency is determined by the slowest layer (i.e., the layer handling the largest set of inputs). In the architectures detailed in \autoref{tab:exp_setup}, the slowest layer is the input layer, processing 784 inputs for MNIST and 700 for \gls{SHD}. In this layer, 100\% of time-steps contain at least one spike for MNIST, while for \gls{SHD}, the percentage is around 48\%. Consequently, \gls{SHD}, with the combination of a lower number of inputs and lower activity in the input layer, enables increased inference speed.
		Since both models use the same number of time steps and clock frequency, and the difference in the number of inputs is not significant, one might expect about 48\% inference time reduction for \gls{SHD} compared to MNIST due to the reduced activity (i.e., about 0.37 ms). However, the observed value in \autoref{tab:SHDbenchmark} is 0.54 ms. The higher latency is explained by the \gls{fcr} model used by \gls{SHD} incorporating inter-layer feedback connections, processed sequentially. Therefore, an additional set of 200 feedback inputs must be processed for the first layer, with an average of 93\% active time steps.

\begin{figure}[hbt!]
				\centering
				\begin{subfigure}[c]{\columnwidth}
					\includegraphics[width=\textwidth]{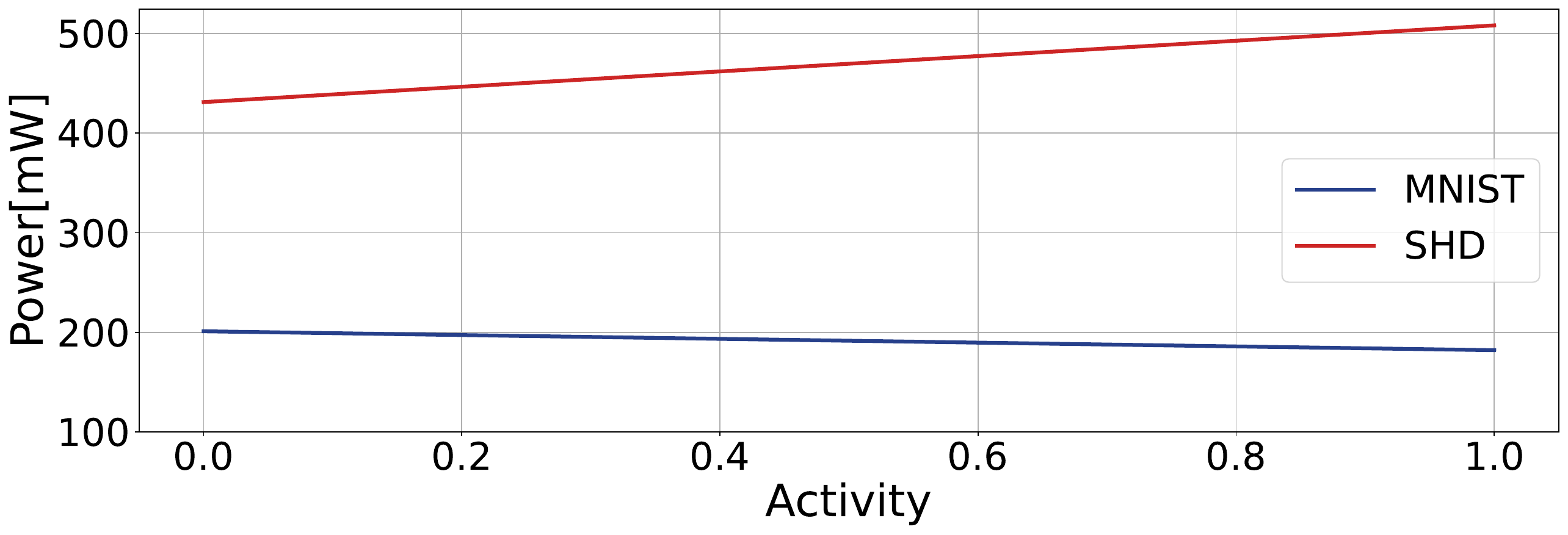}
					\caption{Power}
					\label{fig:pow_vs_act}
				\end{subfigure}
				\vfill
				\begin{subfigure}[c]{\columnwidth}
					\includegraphics[width=\textwidth]{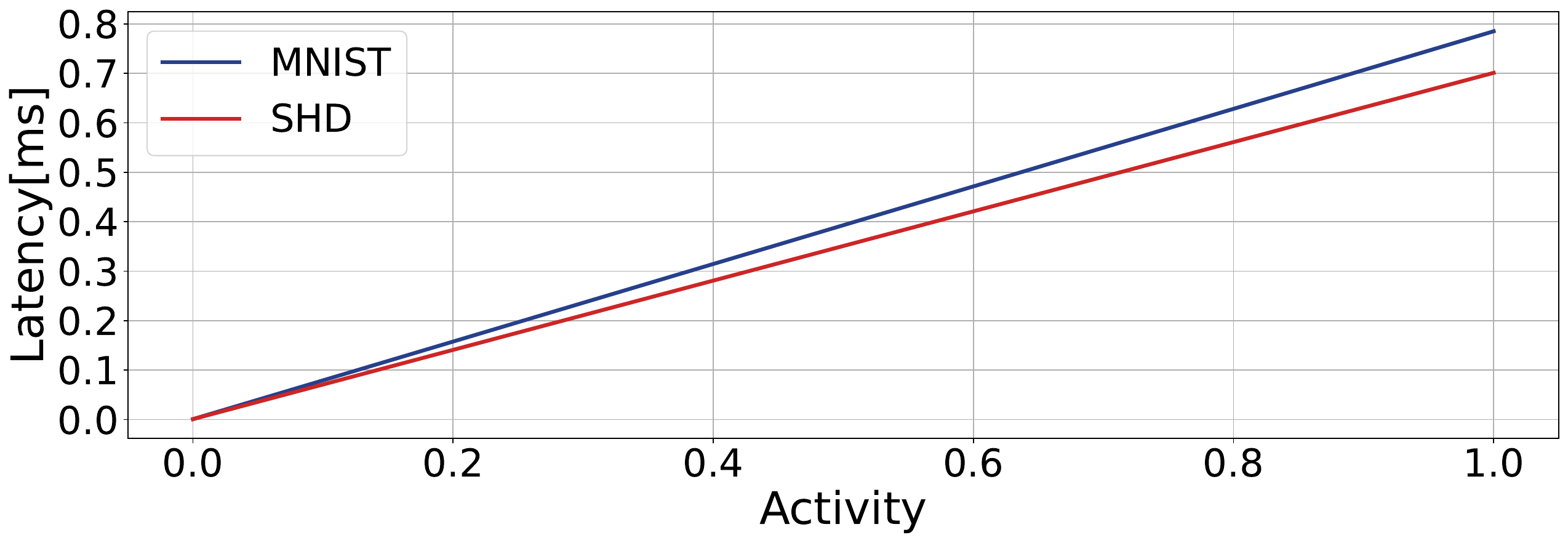}
					\caption{Latency}
					\label{fig:lat_vs_act}
				\end{subfigure}
				\begin{subfigure}[c]{\columnwidth}
					\includegraphics[width=\textwidth]{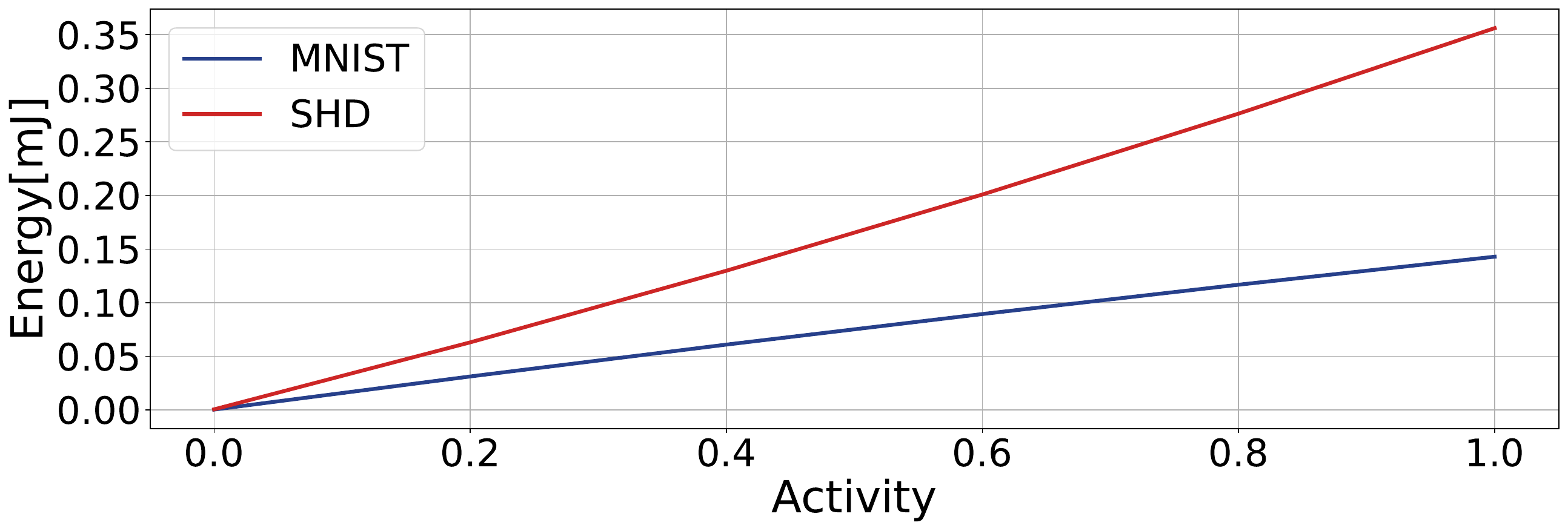}
					\caption{Energy}
					\label{fig:en_vs_act}
				\end{subfigure}
				\vfill
				\caption{Impact of input activity variations on energy, power, and latency of inference using \spikerframework{}}
				\label{fig:pow_lat_en_vs_act}
			\end{figure}

		The input activity not only impacts inference latency, as explained earlier. When the activity decreases, there is a higher probability of having spare cycles with no spikes, allowing the Layer \gls{CU} to skip the sequential processing of all inputs. This reduction in calculations also affects power consumption. \autoref{fig:pow_lat_en_vs_act} analyzes how power, latency, and energy (i.e., power times latency) change with different input activities on the two datasets, highlighting counterintuitive behaviors.

		Examining \autoref{fig:pow_vs_act} that reports the average power consumption of an inference task, we observe two distinct behaviors for MNIST and \gls{SHD}. In the former case, reducing input activity increases power consumption, reaching a limit of around 200mW as activity approaches zero and stabilizing at 180mW under full activity. This behavior is explained by the clock-driven nature of the accelerator, where every clock cycle triggers a network update regardless of active spikes in the input. As mentioned earlier, \spikerframework{} skips time steps without active inputs. However, the exponential decay of the membrane is computed step by step. Therefore, without input stimuli, \spikerframework{} dedicates one clock cycle to decay all membranes before returning to an idle state, awaiting the next input set. The two situations are similar from the perspective of neuron switching power, as the membrane is updated in both cases.
		However, the layer \gls{CU} continually switches between two states, resulting in a high total switching activity and higher power consumption. Conversely, when executing a sequential update on the inputs, the \gls{CU} enters the update state and then awaits the completion of the loop. This behavior is not observed in \gls{SHD}. Given the larger architecture and higher weight bit-width, power consumption in \gls{SHD} is likely dominated by \glspl{BRAM} access during input processing.

		In conclusion, latency heavily depends on input activity. Without active spikes, the execution time tends to  $\texttt{N\_cycles} \times \mathrm{T_{clk}}$, as a single clock cycle is sufficient to decay the membranes. Consequently, overall energy consumption is reduced with decreasing input activity.

		\subsection{Performance vs quantization}
		\label{subsec:quantization}

\begin{figure}[hbt!]
				\centering
				\begin{subfigure}[c]{\columnwidth}
					\includegraphics[width=\textwidth]{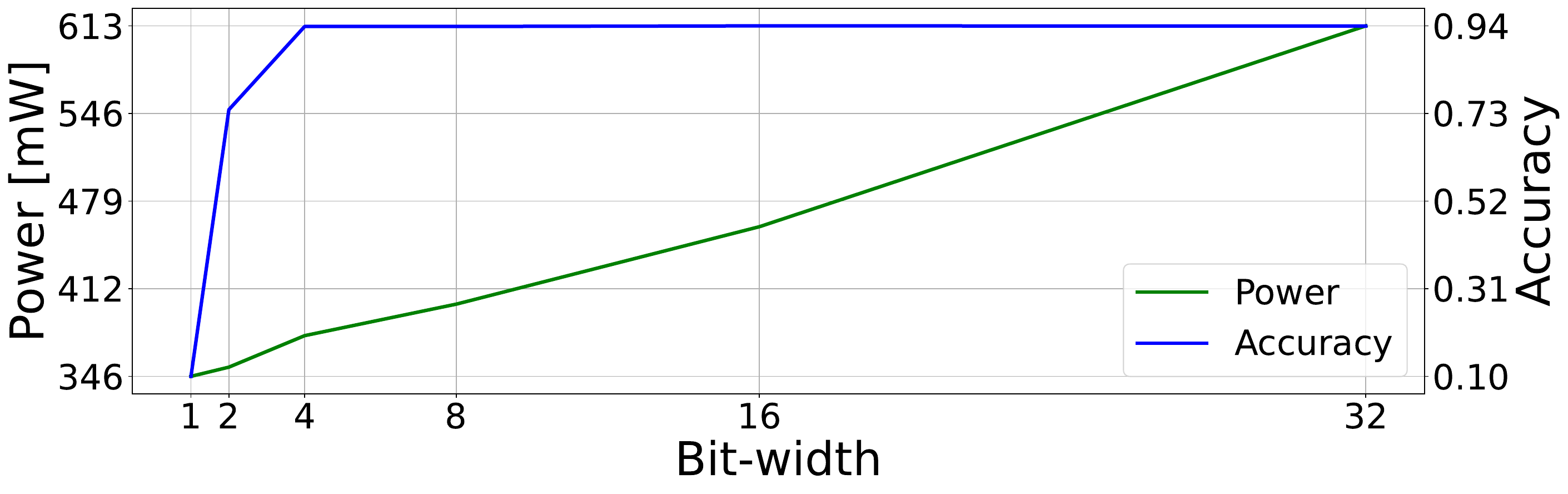}
					\caption{Membrane potential quantization}
					\label{fig:mnist_weights_quant}
				\end{subfigure}
				\hfill
				\begin{subfigure}[c]{\columnwidth}
					\includegraphics[width=\textwidth]{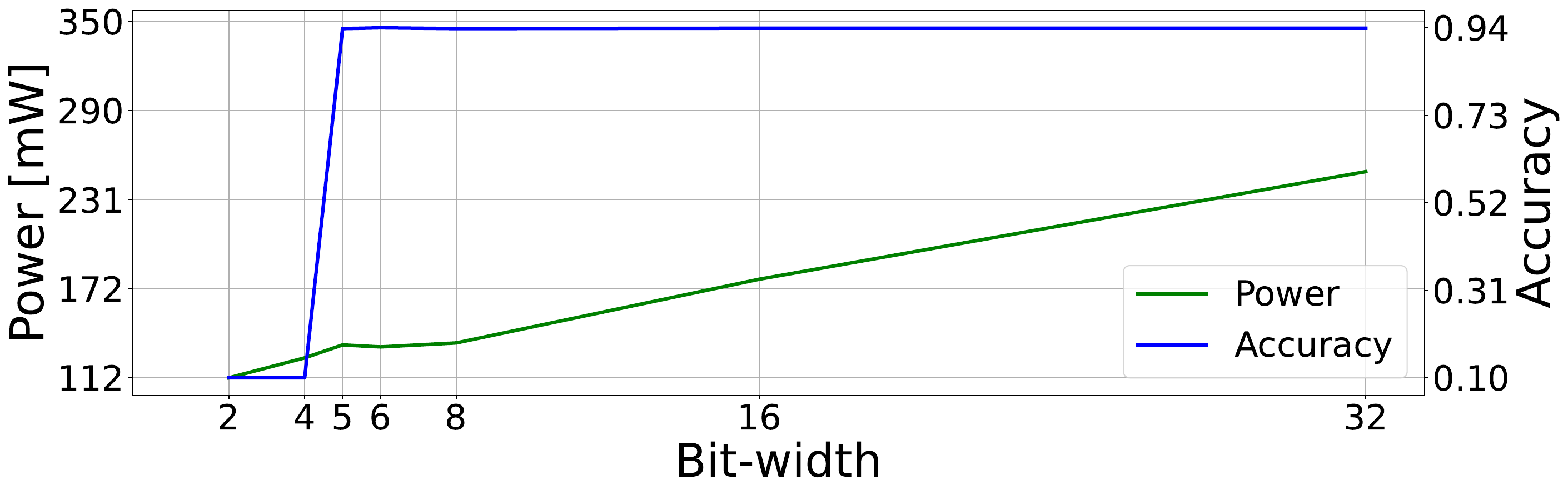}
					\caption{Synaptic weights quantization}
					\label{fig:mnist_neuron_quant}
				\end{subfigure}

				\caption{Impact of quantization of neuron membrane potentials (\ref{fig:mnist_neuron_quant}) and synaptic weights (\ref{fig:mnist_weights_quant}) on inference accuracy and power consumption for MNIST}
				\label{fig:mnist_quant}
			\end{figure}

\begin{figure}[hbt!]
				\centering
				\begin{subfigure}[c]{\columnwidth}
					\includegraphics[width=\textwidth]{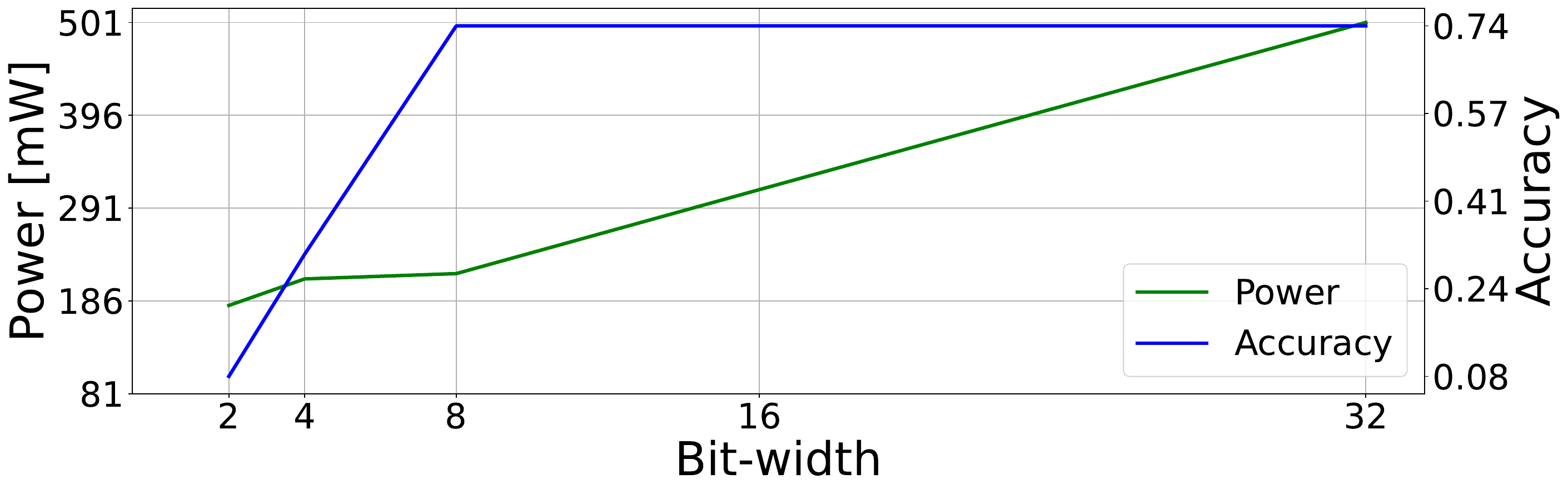}
					\caption{Membrane potential quantization}
					\label{fig:shd_neuron_quant}
				\end{subfigure}
				\hfill
				\begin{subfigure}[c]{\columnwidth}
					\includegraphics[width=\textwidth]{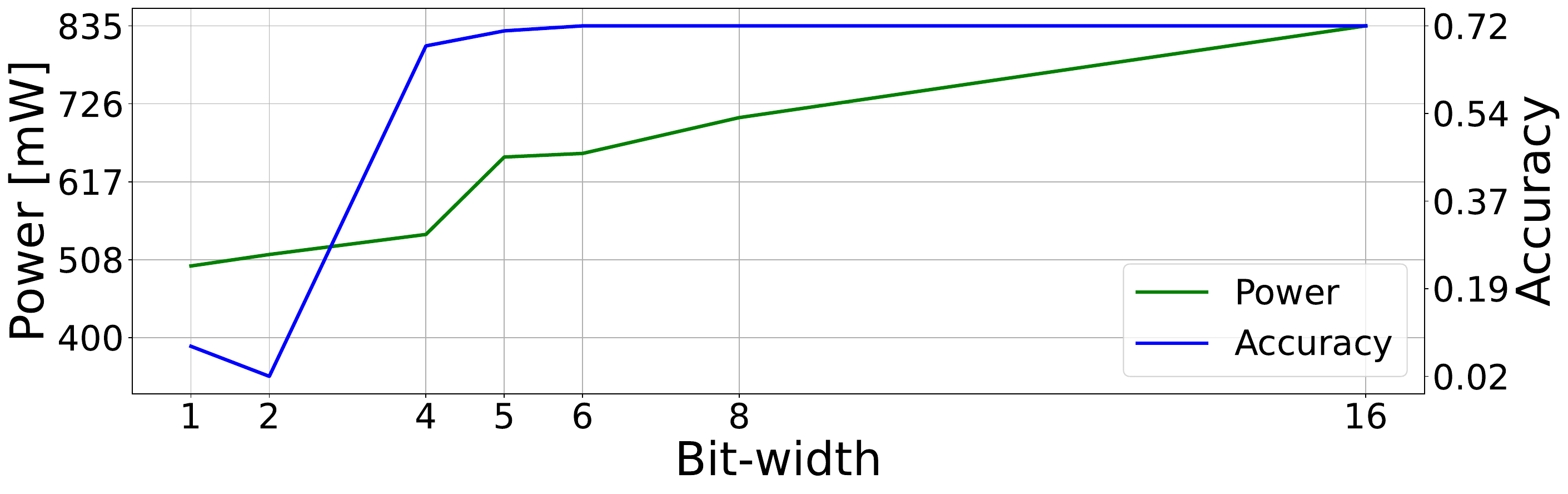}
					\caption{Feed-forward synaptic weights quantization}
					\label{fig:shd_w_ff_quant}
				\end{subfigure}
				\hfill
				\begin{subfigure}[c]{\columnwidth}
					\includegraphics[width=\textwidth]{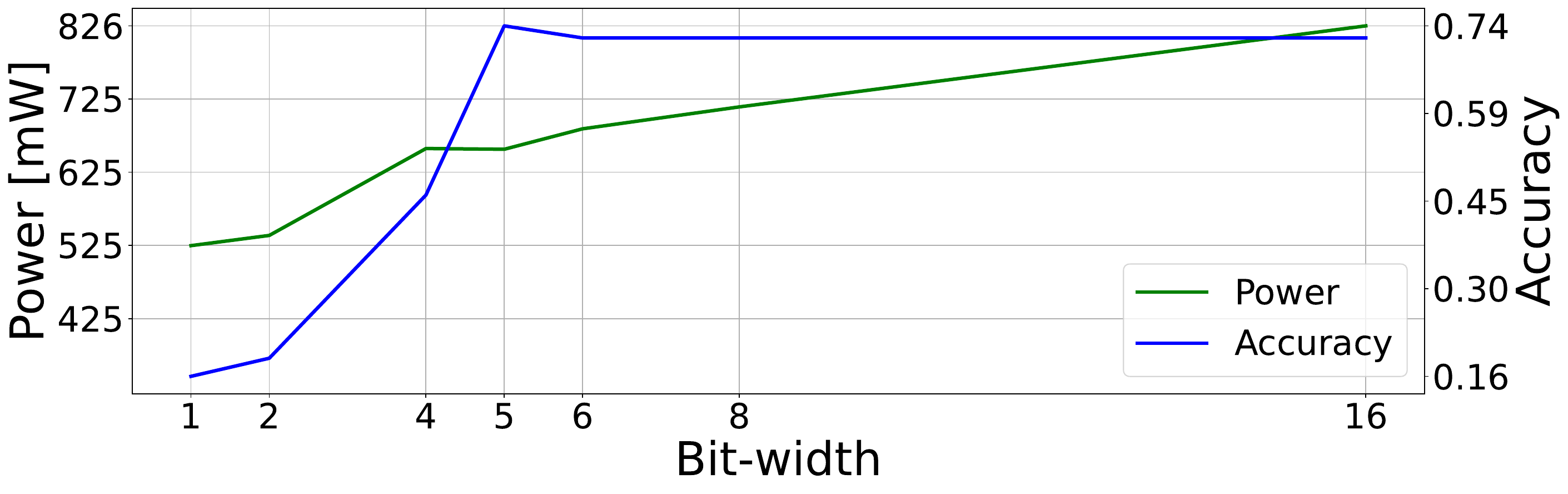}
					\caption{Feedback synaptic weights quantization}
					\label{fig:shd_w_fb_quant}
				\end{subfigure}

				\caption{Impact of quantization of neuron membrane potentials (\ref{fig:shd_neuron_quant}), feed-forward (\ref{fig:shd_w_ff_quant}) and feedback synaptic weights (\ref{fig:shd_w_fb_quant}) on inference accuracy and power consumption for \gls{SHD}}

				\label{fig:shd_quant}
			\end{figure}

		As one of the key features of \spikerframework{} is the optimization of the accelerator through quantization of weights and membrane potentials, it is crucial to examine how these design choices influence performance. Latency is not expected to change, as it is independent of the chosen bit-widths. The primary presumed impacts are on power consumption and network accuracy. Figures \ref{fig:mnist_quant} and \ref{fig:shd_quant} illustrate the results of quantization on MNIST and \gls{SHD}, respectively.

		The resilience of \glspl{snn} to quantization is notable. In both models, despite their differences and the unique information encoding methods employed, the decrease in accuracy with different quantization values is minimal. However, reducing precision by even a single bit has an evident effect on power consumption, owing to the large number of units working in parallel.

		Finally, as explained in \autoref{subsec:optimal_sizing}, it is crucial to note that the primary constraint on \spikerframework{} size is the number of weights that can be accessed in parallel. A smaller weight bit-width reduces the required interconnections and the amount of memory used, including the total number of \glspl{BRAM}. This not only conserves power but also facilitates the implementation of larger architectures.

		\subsection{Performance vs sizing}
		\label{subsec:optimal_sizing}

		Finally, let us explore the model complexity achievable with \spikerframework{} on selected Xilinx\textsuperscript{\texttrademark} FPGA boards. Synthesis results for three Xilinx\textsuperscript{\texttrademark} boards, particularly low-end ones suitable for resource-constrained edge applications, are presented in \autoref{tab:max_size_snn}.

		On the Xilinx\textsuperscript{\texttrademark} XC7Z020/XA7Z020 boards discussed in \autoref{subsec:benchmarking}, the largest \gls{fffc} network possible using a I-order \gls{lif} consists of 1,220 neurons, utilizing 138 \glspl{BRAM} (98.5\%) and 42,430 \glspl{LUT} (26.7\%). The \gls{BRAM} size of the FPGA emerges as the limiting factor. On the slightly more advanced Xilinx\textsuperscript{\texttrademark} XCZU3EG board, a larger 1,900-neuron architecture, utilizing 215 \glspl{BRAM} (99.5\%) and 62,989 \glspl{LUT} (29.8\%), can be implemented. Notably, the place-and-route algorithm encounters no obstacles, reinforcing that \gls{BRAM} limitations govern the network size.

		For \gls{fcr} architectures using a II-order \gls{lif}, where both feed-forward and feedback weights need storage in \gls{BRAM}, the maximum network size is influenced. Specifically, it is capped at 550 neurons for Xilinx\textsuperscript{\texttrademark} XC7Z020/XA7Z020 boards and 690 neurons for the Xilinx\textsuperscript{\texttrademark} XCZU3EG board.

		Potential issues such as place-and-route complexities or excessive power consumption due to the fully parallel nature of the accelerator are foreseeable. A prospective solution involves implementing a certain degree of time multiplexing to address these challenges while expanding the architecture size. This approach entails sharing hardware components between neurons, introducing a trade-off with performance and is currently under study as an enhancement.

			\begin{table*}[htb!]

				\centering

				\caption{Synthesis of maximum size accelerator on different Xilinx\textsuperscript{\texttrademark} \glspl{fpga} boards}
				\label{tab:max_size_snn}

				\begin{tabular}{|c|c|c|c|c|c|c|c|c|c|c|}

					\hline

					\multicolumn{3}{|c|}{\textbf{Target system}}	& 
					\multicolumn{4}{c|}{\textbf{FFFC}}		& 
					\multicolumn{4}{c|}{\textbf{RSNN}}		\\

					\hline

					\textbf{FPGA}					& 
					\textbf{LUT}					& 
					\textbf{BRAM}					&
					\textbf{Neurons}				& 
					\textbf{BRAM}					&
					\textbf{LUT}					&
					\textbf{Power}					&
					\textbf{Neurons}				& 
					\textbf{BRAM}					&
					\textbf{LUT}					&
					\textbf{Power}					\\

					\hline

					XC7Z020 / XA7Z020				&
					159600						&
					140 (560KB)					&
					1224						&
					138						&
					44330						&
					1.07 W						&
					550		& 
					135		& 
					24,660		& 
					720mW		\\ 

					\hline

					XCZU3EG						&
					211,680						&
					216 (864KB)					&
					1900             			& 
					215              			& 
					62989               		& 
					1.2W			            & 
					690              			& 
					213              			& 
					29,809           			& 
					780mW            			\\ 

					\hline









				\end{tabular}
			\end{table*}
			

	\section{Conclusions}
	\label{sec:conclusions}

	This paper introduced \spikerframework{}, a versatile framework to design low-power and resource-efficient hardware accelerators for \glspl{snn} targeting edge inference on \gls{fpga} platforms. It features a Python configuration framework that facilitates easy reconfiguration of the accelerator, allowing users to choose from six neuron models (\gls{if}, I-order \gls{lif}, and II-order \gls{lif}, each with the option of a \emph{hard} or \emph{subtractive} reset) and two network architectures (\gls{fffc} and \gls{fcr}). The tool enables the automatic selection of training and quantization parameters directly through Python. The results are significant, boasting a 93.85\% accuracy on MNIST, with a classification latency of $780\mu s$ per image and power consumption of $180mW$. Additionally, it achieves a 72.99\% accuracy on \gls{SHD}, corresponding to a $540\mu s$ latency and power consumption of $430mW$. These metrics are highly competitive compared to state-of-the-art \gls{fpga} accelerators for \glspl{snn}, demonstrating high performance in both power efficiency and area. This work lays a solid foundation for deploying specialized, low-power, and efficient \gls{snn} accelerators in resource and power-constrained edge applications. \spikerframework{} is a live project, and ongoing work focuses on enlarging the library of available neurons, input encoding blocs, and network architectures. To encourage research in this field, \spikerframework{} is available as an open-source project\footnote{Link to a public GitHub repo will be released in case of publication of the paper}.

	\bibliographystyle{IEEEtran}
	\bibliography{IEEEabrv, Bibliography/bibliography.bib}

	\vfill

	\begin{IEEEbiography} [{\includegraphics[width=1in,height=1.25in,clip,keepaspectratio]{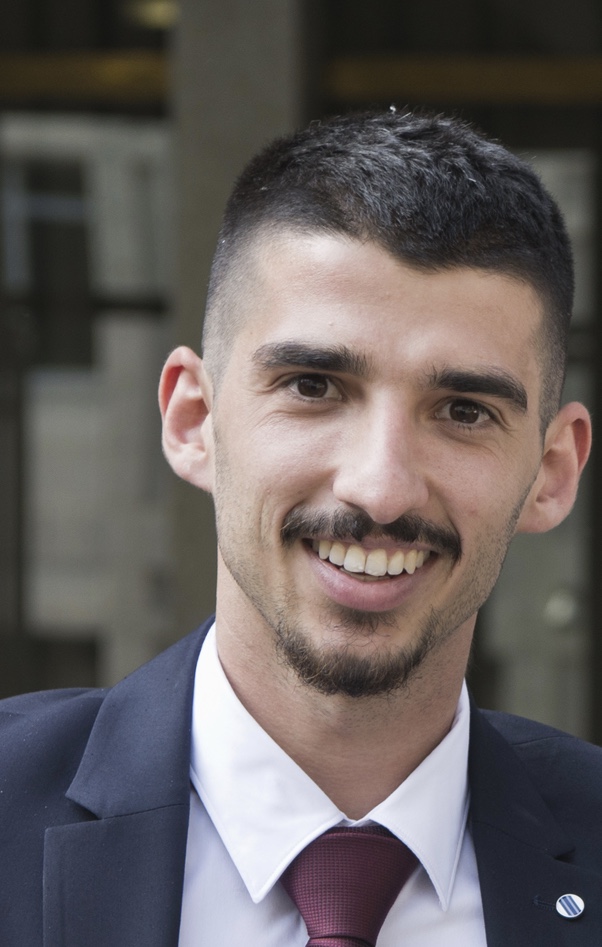}}] {Alessio Carpegna} Alessio (student, IEEE '21) received an M.Sc. degree in Electronic engineering from the Politecnico di Torino, Torino, Italy, in 2021, with a specialization in digital systems design. In the same year he entered the Italian national Ph.D. program in Artificial Intelligence. His current research interests include Neuromorphic Systems, and in particular the design of neuromorphic hardware accelerators, Operating Systems and Embedded Systems in general.
	\end{IEEEbiography}

	\begin{IEEEbiography}[{\includegraphics[width=1in,height=1.25in,clip,keepaspectratio]{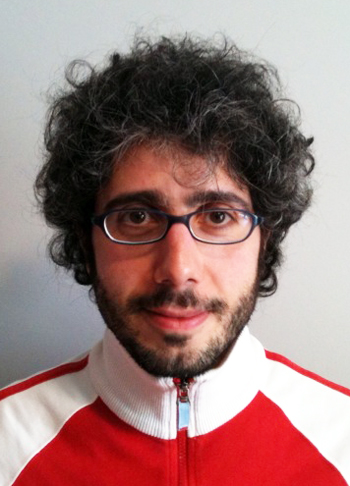}}]{Alessandro Savino}(M'14, SM'22) is an Assistant Professor in the Department of Control and Computer Engineering at Politecnico di Torino (Italy). He holds a Ph.D. (2009) and an M.S. equivalent (2005) in Computer Engineering and Information Technology from the Politecnico di Torino in Italy. Dr. Savino's research contributions include Approximate Computing, Reliability Analysis, Safety-Critical Systems, Software-Based Self-Test, and Image Analysis. He has been part of the program and organizing committee of several IEEE and INSTICC conferences and has served as a reviewer of IEEE conferences and journals. His research interests include Operating Systems, Imaging algorithms, Machine Learning, Evolutionary Algorithms, Graphical User Interface experience, and Audio manipulation.
	\end{IEEEbiography}

	\begin{IEEEbiography}[{\includegraphics[width=1in,height=1.25in,clip,keepaspectratio]{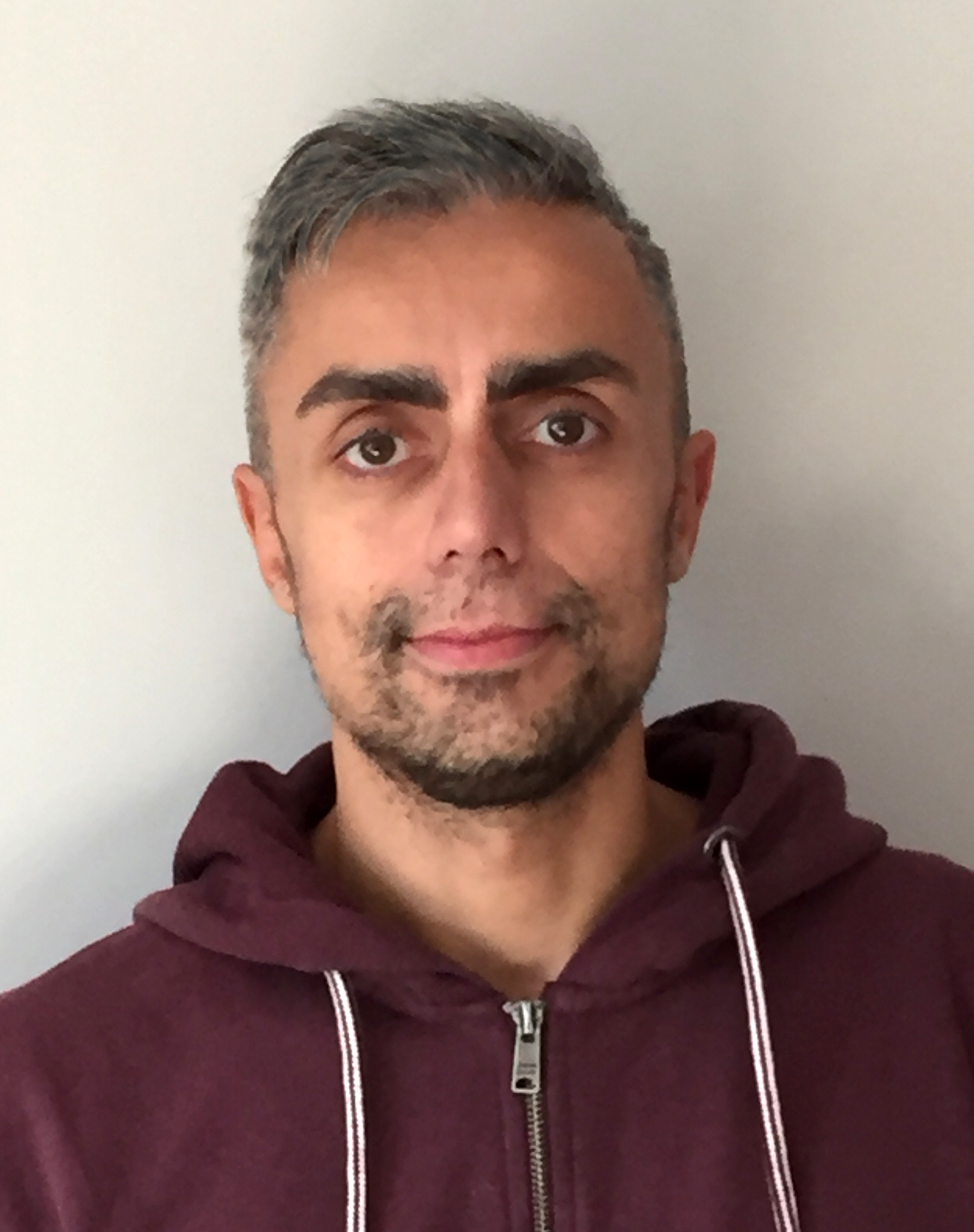}}]{Stefano Di Carlo}
	(SM'00-M'03-SM'11) He received an M.Sc. degree in computer engineering and a Ph.D. in information technologies from Politecnico di Torino, Italy, where he is a Full Professor. His research interests include DFT, BIST, and dependability. He has coordinated several national and European research projects. Di Carlo has published over 200 papers in peer-reviewed IEEE and ACM journals and conferences. He regularly serves on the Organizing and Program Committees of major IEEE and ACM conferences. He is a Golden Core member of the IEEE Computer Society and a senior member of the IEEE.
	\end{IEEEbiography}

\end{document}